\documentclass[10pt,journal,compsoc]{IEEEtran}

\ifCLASSOPTIONcompsoc
  \usepackage[nocompress]{cite}
\else
  \usepackage{cite}
\fi

\ifCLASSINFOpdf
\else
\fi

\usepackage[colorlinks, citecolor=green]{hyperref}
\usepackage{graphicx}
\usepackage{amssymb}
\usepackage{amsmath}
\usepackage{amsfonts}
\usepackage{subfig}
\usepackage{array}
\usepackage{booktabs}
\usepackage{bm}
\usepackage{cuted}
\usepackage{latexsym}
\usepackage{diagbox}
\usepackage[misc]{ifsym}
\usepackage{url}
\usepackage{xcolor}
\usepackage{lineno,hyperref}
\usepackage{makecell}
\begin{document}
%
\title{Gaussian-Hermite Moment Invariants of General Multi-Channel Functions}

\author{Hanlin~Mo,
	    Hua~Li,~\IEEEmembership{Senior Member, IEEE}, and 
        Guoying Zhao,~\IEEEmembership{Fellow, IEEE}
\IEEEcompsocitemizethanks{\IEEEcompsocthanksitem H. Mo is with the Center for Machine Vision and Signal Analysis, University of Oulu, FI-90014, Oulu, Finland. E-mail: hanlin.mo@oulu.fi
\IEEEcompsocthanksitem H. Li is with the Key lab of Intelligent Information Processing, the Institute of Computing Technology, Chinese Academy of Sciences, 100190 Beijing; University of Chinese Academy of Sciences, 100049 Beijing. E-mail: lihua@ict.ac.cn 
\IEEEcompsocthanksitem G. Zhao (corresponding author) is with the Center for Machine Vision and Signal Analysis, University of Oulu, FI-90014, Oulu, Finland, and is also with the School of Information and Technology, Northwest University, Xi'an 710069, China. E-mail: guoying.zhao@oulu.fi}}

\markboth{Journal of \LaTeX\ Class Files}%
{Shell \MakeLowercase{\textit{et al.}}: Bare Demo of IEEEtran.cls for Computer Society Journals}

\IEEEtitleabstractindextext{%
\begin{abstract}
With the development of data acquisition technology, large amounts of multi-channel data are collected and widely used in many fields. Most of them, such as RGB images and vector fields, can be expressed as different types of multi-channel functions. Feature extraction of multi-channel data for identifying interest patterns is a critical but challenging task. This paper focuses on constructing moment-based features of general multi-channel functions. Specifically, we define two transform models, rotation-affine transform and total rotation transform, to describe real deformations of multi-channel data. Then, we design a structural framework to generate Gaussian-Hermite moment invariants for these two transform models systematically. It is the first time that a unified framework has been proposed in the literature to construct orthogonal moment invariants of general multi-channel functions. Given a specific type of multi-channel data, we demonstrate how to utilize the new method to derive all possible invariants and eliminate dependences among them. We obtain independent sets of invariants with low orders and low degrees for RGB images, 2D vector fields and color volume data. Based on synthetic and real multi-channel data, we conduct extensive experiments to evaluate the stability and discriminability of these invariants and their robustness to noise. The results show that new moment invariants significantly outperform previous moment invariants of multi-channel data in RGB image classification and vortex detection in 2D vector fields. 
\end{abstract}

}
\maketitle
\IEEEdisplaynontitleabstractindextext
\IEEEpeerreviewmaketitle

\IEEEraisesectionheading{\section{Introduction}\label{sec:introduction}}

\IEEEPARstart{H}{ow} to extract effective features of various types of data is one of the most fundamental problems in pattern recognition. A desirable feature should be able to capture intrinsic information of a given data, which means it should be invariant to the deformations caused by the sensor's setup, the influence of the environment and many other factors. Researchers have designed numerous invariant features of different data for different practical applications. Among them, moments and moment invariants play a critical role.

From the mathematical point of view, moments are "projections" of a function on a polynomial basis. Classical moment invariants are special polynomials of moments that have invariance to certain transform models. For example, the standard power basis leads to geometric moments, which are widely used in the image analysis community. In 1962, based on the theory of algebraic invariants, Hu first derived seven geometric moment invariants of grayscale images to 2D rotations \cite{1}. Thirty years later, Reiss, Flusser and Suk independently published several affine moment invariants \cite{2,3,4}. Since then, an effort has been put into designing simpler and more transparent ways to generate rotation or affine moment invariants of grayscale images systematically \cite{5,6}. In 2018, Li et~al. further proved the existence of image projective invariants using finite combinations of weighted geometric moments \cite{7}. Unfortunately, geometric moments and moment invariants are very sensitive to additive noise and have a high level of information redundancy. To solve these problems, a series of papers focused on defining image moments using various orthogonal polynomial bases. The experimental results in these papers clearly showed that orthogonal moments have better numerical stability and recognition ability than geometric moments and complex moments. We can divide the existing orthogonal moments into two groups: moments orthogonal on a circle and moments orthogonal on a square. The former group includes Zernike moments \cite{8,9}, Pseudo Zernike moments \cite{10}, Fourier-Mellin moments \cite{11,12}, Jacobi-Fourier moments \cite{13} and Chebyshev-Fourier moments\cite{14}. The values of these moments are complex numbers, and their magnitudes are naturally invariant to 2D rotations. Before calculating them, we must map a grayscale image into a unit circle. The interpolation operation not only leads to the loss of precision but also increases the computational time. Hence, the latter one, moments orthogonal on a square, are preferred by many researchers, such as Legendre moments \cite{15,16}, Chebyshev moments \cite{17,18}, Krawtchouk moments \cite{19,20}, Gegenbauer moments \cite{21} and Gaussian-Hermite moments \cite{22}. However, constructing rotation invariants from these moments takes a lot of work. The only exception is the Gaussian-Hermite moments. Yang et al. proved that any geometric moment invariants to 2D rotations remain invariant when replacing geometric moments with the corresponding Gaussian–Hermite moments. Based on this discovery, they derived a set of Gaussian-Hermite moment invariants to 2D rotations \cite{23,24} and further achieved their invariance to 2D scale transform \cite{25}.   

\begin{figure}
	\centering
	\subfloat[RGB images.]
	{\includegraphics[height=10mm,width=28mm]{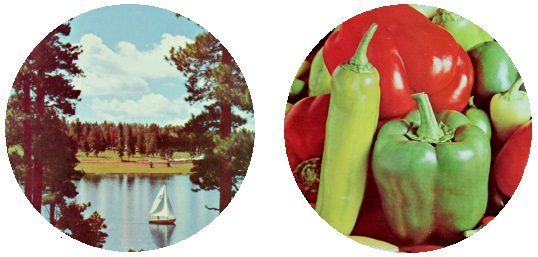}\label{figure:1(a)}\hfill}~~~~
	\subfloat[2D von K$\acute{a}$rm$\acute{a}$n vortex street.]
	{\includegraphics[height=10mm,width=50mm]{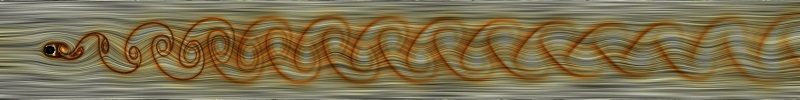}\label{figure:1(b)}\hfill}\\
	\subfloat[3D vector field around a half cylinder and color volume data composed of a series of RGB cyosections. As shown in Table. \ref{tab1e:1}, both of them can be expressed as the same type of multi-channel functions.]
	{\includegraphics[height=13mm,width=80mm]{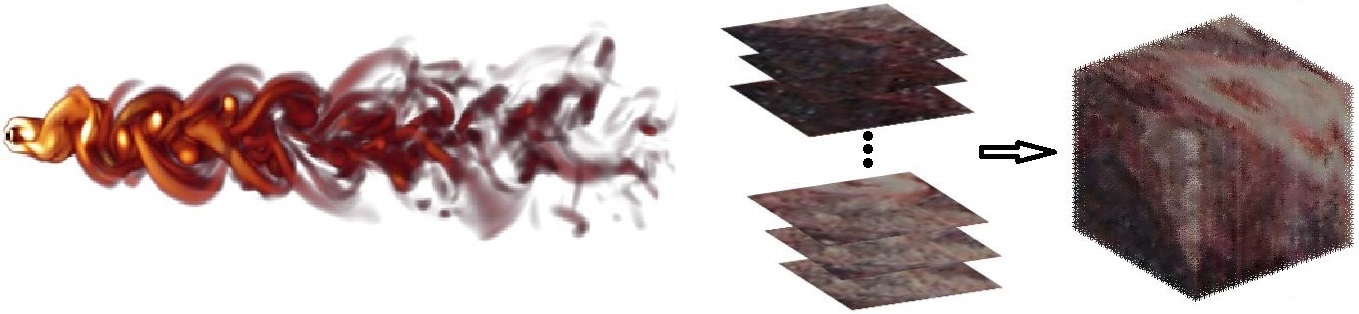}\label{figure:1(c)}\hfill}~~~~
	\caption{Several common instances of multi-channel data.}\label{figure:1}
\end{figure}

All of the above-mentioned moment invariants are calculated from grayscale images. Some construction methods of these invariants can also be generalized to derive moment invariants of 3D shapes \cite{26,27}. Both grayscale images and 3D shapes are single-channel data. We can express them as different scalar functions. When constructing moment invariants of scalar functions, researchers just discussed their invariance to geometric transforms acting on spatial coordinates. With the development of simulation and measurement techniques, multi-channel datasets are rapidly growing in size and becoming prevalent in many fields, including pattern recognition, computer vision, visualization and computer graphics. These multi-channel data, such as RGB images, 2D vector fields, 3D vector fields and color volume data (see in Fig.\ref{figure:1}), can be regarded as different types of multi-channel functions. In most cases, traditional geometric transforms cannot describe realistic deformations of multi-channel functions because these deformations usually simultaneously act on spatial coordinates and function values. To address this issue, researchers designed several more complicated and effective transform models for multi-channel data, such as total rotation transform $(TR)$ and total affine transform $(TA)$. In the last decade, moment invariants of multi-channel functions to these transform models have attracted considerable attention. According to data types, we can classify relevant moment invariants into the following three categories

\textbf{RGB images}: There are a lot of papers about constructing moment invariants of color images, but most of them are just invariant to geometric transforms \cite{28,29,30,31,32,33}. In 2017, using geometric primitive introduced by Xu and Li \cite{26}, Gong et~al. first derived geometric moment invariants of RGB images to $TA$ \cite{34}. These features are invariant to both the 2D affine transform of spatial coordinates and the 3D affine transform of color space. In fact, $TA$ is the best linear model to simulate the deformations of color images caused by imaging geometry and the changes of illumination condition \cite{2,3,35,36,37}. Before that, Mindru et al. also constructed similar invariants under restricted $TA$, in which the color affine transform degenerated into a 3D diagonal transform \cite{38,39}.    

\textbf{2D vector fields}: In 2017, Schlemmer et al. first generated complex moment invariants of 2D vector fields to $TR$ \cite{40,41}. Then, Bujack et al. obtained invariant complex moments to $TR$ by using a normalization approach \cite{42,43,44,45}. Specifically, they estimated the parameters of $TR$ using several complex moments with low orders. However, using this approach, we can not accurately determine the angle of rotation, which is the most important parameter of $TR$, and have to calculate normalized complex moments for a set of candidate angles. In addition, the potential errors of transform parameters will affect the stability of all invariant moments. However, this issue has yet to be studied and experimentally tested in their papers \cite{46}. In 2018, Yang et al. constructed orthogonal moment invariants to $TR$ from Zernike moments and Gaussian-Hermite moments \cite{47,48}. Recently, Kostkov$\acute{a}$ et al. also designed a procedure to systematically generate geometric moment invariants to $TA$ \cite{49,50}, which is similar to Gong's approach \cite{34}.        

\textbf{3D vector fields and color volume data}: Since most of the fluid flow data are intrinsically three-dimensional, researchers are more desirable to get moment invariants of 3D vector fields than those of 2D ones. Unfortunately, there needs to be more work on this topic. The main reason is that we can only extend a few methods of deriving moment invariants of 2D vector fields to 3D. For 3D vector fields and even more general tensor fields, Hagen and Langbein designed tensor-valued functions of geometric moments, which are invariant to $TR$ \cite{51}. However, these moment tensors are not classical moment invariants, and the paper \cite{51} neither provided explicit formulas of them nor carried out numerical experiments to test their performance. Based on their previous work, Bujack et al. proposed a more complicated normalization approach to calculate invariant geometric moments of 3D vector fields to $TR$ \cite{52}. Recently, Hao et~al. generalized Gong's method \cite{34} to generate geometric moment invariants of general multi-channel functions to $TA$ \cite{53}, and first showed the expansions of classical moment invariants of color volume data. We can regard color volume data and 3D vector fields as the same type of multi-channel data.     

There are three major limitations of the methods above. \textbf{(1)} Most of them use geometric moments or complex moments to construct moment invariants of multi-channel functions. As mentioned previously, they are not robust to noise, which substantially limits their usage in practical tasks. \textbf{(2)} Many methods can not be extended to handle higher-dimensional multi-channel data. For example, Yang et al. represented a 2D vector field as a complex-valued function. Based on this representation and the isomorphism between Gaussian-Hermite moments and geometric moments under 2D rotations, they derived Gaussian-Hermite moment invariants of 2D vector fields to $TR$ \cite{48}. However, there is no concept corresponding to the complex number in higher-dimensional space, which means that we can not similarly represent 3D, 4D or 5D vector fields. Meanwhile, as shown in \cite{27}, it is challenging to prove that the isomorphism relationship between two kinds of moments still holds for higher-dimensional data. As a result, Yang et al. did not get orthogonal moment invariants of 3D vector fields to $TR$ using this approach. \textbf{(3)} Some methods are just suitable for vector fields instead of general multi-channel functions. In other words, these methods demand the channel dimension of a given multi-channel function must be equal to its coordinate dimension \cite{40,42,48,52}. Thus, they can not be applied to handle many widely used multi-channel data which do not meet this condition, such as RGB images.   

The goal of our paper is to address these limitations, and the main contributions can be summarized as follows
\begin{itemize}
	\item We define two transform models of multi-channel functions, rotation-affine transform $(RA)$ and total rotation transform $(TR)$, and give some specific instances of them. From a practical point of view, we also explain why $RA$ and $TR$ can be used to describe realistic deformations of vector fields.     
	\item Using two fundamental differential operators and two fundamental primitives, we develop a novel method to derive orthogonal Gaussian-Hermite moment invariants of general multi-channel functions to $RA$ and $TR$, which are denoted as $MGHMIs$. 
	\item For RGB images, 2D vector fields and color volume data, all possible $MGHMIs$ with low orders and low degrees are generated. We identify the dependencies among them and derive independent sets. 
	\item Numerical experiments are carried out on synthetic and real multi-channel data. We demonstrate the stability and discriminability of $MGHMIs$ and test their robustness to additive noise. Some current moment invariants of multi-channel data are chosen for comparison. The experimental results show that $MGHMIs$ have better performance in RGB image classification and vortex detection in 2D vector fields.    
\end{itemize} 
The rest of our paper is organized as follows. Section \ref{section:2} formulates the definitions and concepts used throughout the paper. Sections \ref{section:3} and \ref{section:4} are the main contributions of this paper. We introduce the structural framework of $MGHMIs$ and then derive some instances with low degrees and low orders for widely used multi-channel data. In Section \ref{section:5}, we conduct experiments to validate the performance of $MGHMIs$ in various recognition tasks. Finally, Section \ref{section:6} presents our conclusions.

\section{Basic Definitions and Notations}
\label{section:2}
This section introduces some basic concepts and definitions used in the following sections.

\subsection{Multi-Channel Functions}
\label{section:2.1}
A general multi-channel function $F(X):\Omega\subset\mathbb{R}^{M}\rightarrow\mathbb{R}^{N}$ can be defined as
\begin{equation}\label{equ:1}
	\begin{split}
		&X=(x_{1}, x_{2}, \cdots, x_{M})^{T}\\&
		F(X)=\left(f_{1}(X), f_{2}(X), \cdots, f_{N}(X)\right)^{T}
	\end{split}
\end{equation}
where
\begin{itemize}
	\item Both $M$ and $N$ are positive integers.
	\item For any $n\in\{1,2,\cdots,N\}$, the scalar function $f_{n}(X)$ acts as mapping from the domain $\Omega$ to $\mathbb{R}$. When $N=1$, $F(X)=f_{1}(X)$ is a scalar function. 
\end{itemize}

We can instantiate (\ref{equ:1}) by setting specific $M$ and $N$. Some instances are listed in Table \ref{tab1e:1}. For example, various vector functions, such as 2D and 3D vector fields, can be regarded as a special class of multi-channel functions. For them, we always have $M=N$, i.e. the dimension of the function value $F(X)$ must be equal to the dimension of the spatial coordinate $X$.

\begin{table}
	\caption{\label{tab1e:1}. Some specific instances of (\ref{equ:1}).}
	\centering
	\begin{tabular}{p{1.0cm}p{6.0cm}}
		\toprule[1.1pt]
		\textbf{(M,N)}&\textbf{Instances}\\
		\toprule[1.1pt]
		$(2,3)$ & RGB Images: \newline$F(X)=(R(x,y),G(x,y),B(x,y))$\\
		\midrule
		$(2,2)$& 2D Vector Fields:\newline$F(X)=(U(x,y),V(x,y))$\\
		\midrule
		$(3,3)$ & Color Volume Data: \newline$F(X)=(R(x,y,z),G(x,y,z),B(x,y,z))$ \newline 3D Vector Fields: \newline $F(X)=(U(x,y,z),V(x,y,z),W(x,y,z))$\\
		\midrule
	\end{tabular}
\end{table}

\begin{figure}
	\centering
	\subfloat[$RA$ versions of the same RGB image $(M=2,N=3)$.]
	{\includegraphics[height=13mm,width=80mm]{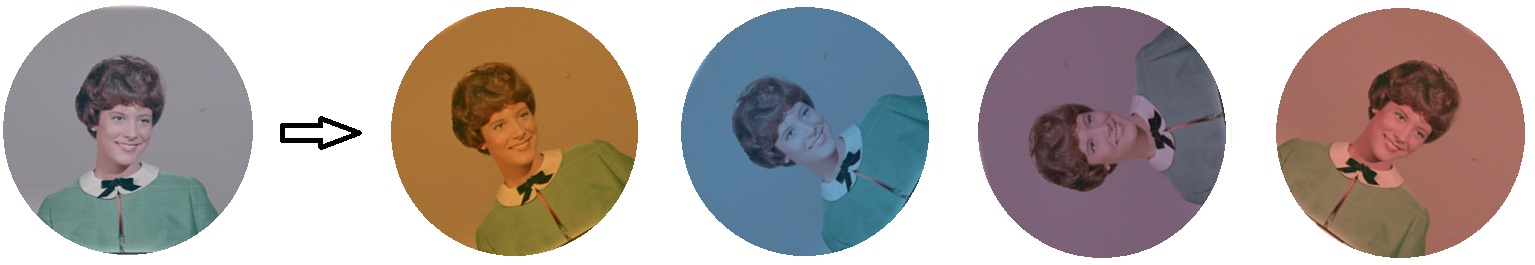}\label{figure:2(a)}\hfill}\\
	\subfloat[Special $TR$ versions of a vector field $(M=2,N=2)$.]
	{\includegraphics[height=13mm,width=80mm]{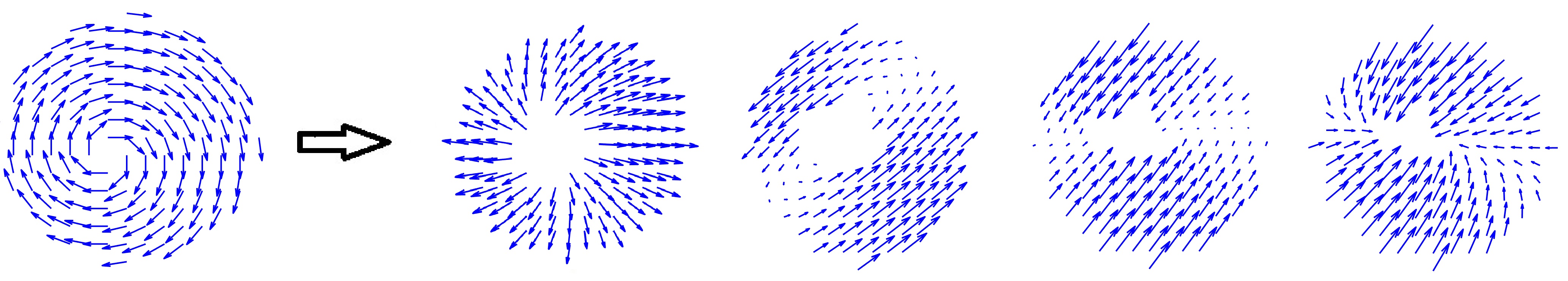}\label{figure:2(b)}\hfill}
	\caption{$RA$ and special $TR$ versions of commonly used multi-channel data.}\label{figure:2}
\end{figure}

\subsection{Transform Models of Multi-Channel Functions}
\label{section:2.2}
In this paper, we mainly discuss two transform models of multi-channel data, $RA$ and $TR$, where $TR$ is just a special case of $RA$. 
\\
\\
\noindent
\textbf{The definitions of \bm{$RA$} and \bm{$TR$}:} If a multi-channel function $F(X)$ defined by (\ref{equ:1}) is changed to $G(Y):\Omega^{'}\subset\mathbb{R}^{M}\rightarrow\mathbb{R}^{N}$ using a rotation-affine transformation $\left(R_{in}; A_{out}, T_{out}\right)$, we have
\begin{equation}\label{equ:2}
	Y=R_{in} \cdot X,~~~~G(Y)=A_{out} \cdot F(X)+T_{out}
\end{equation}
where
\begin{itemize}
	\item $Y$=$(y_{1},\cdots,y_{M})^{T}$ and $G(Y)$=$\left(g_{1}(Y),\cdots, g_{N}(Y)\right)^{T}$.
	\item The inner transformation $IT=R_{in}\in\mathbb{R}^{M\times M}$ acts on $M$-dimensional spatial coordinate $X$. It is a rotation matrix, meaning that $R_{in}^{-1}=R_{in}^{T}$ and the determinant $|R_{in}|=1$.  
	\item The outer transformation $OT=(A_{out},T_{out})$ is an affine transform which acts on $N$-dimensional function value $F(X)$. Note that $A_{out}\in\mathbb{R}^{N\times N}$ is a non-singular matrix and $T_{out}\in \mathbb{R}^{N}$ represents an outer translation.
	\item When the non-singular matrix $A_{out}$ in $OT$ degenerates into a rotation matrix $R_{out}\in\mathbb{R}^{N\times N}$, $RA$ degenerates into total rotation transform ($TR$). It is obvious that we have $TR\subset RA$. 
\end{itemize} 
When constructing invariant features of multi-channel data to $OT$, researchers usually omit the outer translation $T_{out}$ because its parameters can be easily removed by subtracting the average of $F(X)$ over the domain $\Omega$. Thus, for $OT$, our paper will just achieve the invariance to the matrix $A_{out}$. Also, we do not define the inner translation $S_{in}\in \mathbb{R}^{M}$ in $IT$. In fact, when extracting features from a local region around a given point, a local coordinate system is first established with this point as the origin. This makes these features naturally invariant to $S_{in}$.
\begin{figure*}
	\normalsize
	\setcounter{equation}{2}
	\begin{equation}\label{equ:3}
		\left(
		\begin{array}{cc}
			u\\
			v\\
		\end{array}
		\right)
		=
		\left(
		\begin{array}{cc}
			cos(\theta) & -sin(\theta)\\
			sin(\theta) & cos(\theta)\\
		\end{array}
		\right)
		\cdot
		\left(
		\begin{array}{c}
			x\\
			y\\
		\end{array}
		\right)
		+
		\left(
		\begin{array}{c}
			t_{1}\\
			t_{2}\\
		\end{array}
		\right)
		,~~~~
		\left(
		\begin{array}{c}
			R^{'}(u,v)\\
			G^{'}(u,v)\\
			B^{'}(u,v)\\
		\end{array}
		\right)
		=
		\left(
		\begin{array}{ccc}
			a_{11} & a_{12} & a_{13}\\
			a_{21} & a_{22} & a_{23}\\
			a_{31} & a_{32} & a_{33}\\
		\end{array}
		\right)
		\left(
		\begin{array}{c}
			R(x,y)\\
			G(x,y)\\
			B(x,y)\\
		\end{array}
		\right)
		+
		\left(
		\begin{array}{c}
			t_{1}\\
			t_{2}\\
			t_{3}\\
		\end{array}
		\right)
	\end{equation}
\end{figure*}
\\
\\
\noindent
\textbf{The applications of \bm{$RA$} and \bm{$TR$}:}
Since $RA$ and $TR$ simultaneously act on $X$ and $F(X)$, they can model some realistic deformations of widely used multi-channel data.  

Previous studies proved that the 3D affine transform is the best linear model to describe the photometric change of RGB images \cite{35,36,37}. Two RGB images $(R(x,y),G(x,y),B(x,y))$ and $(R^{'}(u,v),G^{'}(u,v),B^{'}(u,v))$ of the same planar object taken from different angles and illumination conditions can be related by (\ref{equ:3}), where the rotation angle $\theta\in (0,2\pi]$, 3D matrix acting on RGB values is non-singular and all parameters are real numbers. We can use (\ref{equ:2}) to represent this transformation via setting $M=2$ and $N=3$. Fig.\ref{figure:2(a)} shows $RA$ versions of an RGB image. Similarly, when setting $M=N=3$, $RA$ can model spatial rotation and color variation of color volume data. When a 2D scalar function $f(x,y)$ is rotated by $\theta$ and $g(u,v)$ denotes the rotated version, their second-order partial derivatives $(f_{xx},f_{xy},f_{yy})^{T}$ and $(g_{uu},g_{uv},g_{vv})^{T}$ are also related by $RA$.   

Vector fields are widely used multi-channel data. As mentioned in Section \ref{section:2.1}, when $F(X)$ is a vector field, we always have $M=N$. In theory, if we transform the spatial coordinate $X$ of a vector field, its vector value $F(X)$ must be transformed using the same transformation, which means $IT\equiv OT$. Based on this property, $RA$ seems unsuitable to model local deformations of vector fields because its $IT$ is a spatial rotation and $OT$ is an affine transform. $TR$ can be used by further limiting $R_{out}=R_{in}$. We call this restricted transform model special $TR$. Fig.\ref{figure:2(b)} shows several special $TR$ versions of a 2D vector field. All "$TR$" appearing in previous papers refers to the special $TR$, and researchers only discussed moment invariants of vector fields to special $TR$ instead of general $TR$ \cite{41,45,48,52}. 

However, in practice, most collected vector field data are imperfect because vector values will be disturbed by many factors, such as calibration error and random noise. Thus, $OT$ is not exactly equal to $IT$. Meanwhile, when detecting similar local patterns in a vector field, such as vortices, these patterns are not transformed versions of the same template. They are related by more complicated transform models which do not meet the constraint $OT=IT$. Unfortunately, moment invariants to special $TR$ are sensitive to small differences between $OT$ and $IT$. As a result, their stability and discriminability will drastically degrade in practical tasks. 

To handle this problem, we generate moment invariants of vector fields to $RA$ and $TR$ in this paper. First, due to special $TR\subset TR\subset RA$, these moment invariants are also invariant to special $TR$. Secondly, since strict constraints do not link $OT$ and $IT$ in $RA$ and $TR$, these invariants are more robust to those disturbances acting on vector values. To verify this, in Section \ref{section:5.2}, we conduct vortex detection experiments on 2D vector fields and test the performance of Gaussian-Hermite moment invariants to special $TR$, $TR$ and $RA$, respectively. The results show that the performance of moment invariants to $RA$ and $TR$ significantly outperforms those only invariant to special $TR$. In particular, moment invariants to $RA$ achieve the best detection results. Therefore, it is meaningful to construct moment invariants of vector field data to $RA$ and general $TR$.
\\
\\         
\noindent
\textbf{Why we do not use \bm{$TA$}:} In $TA$, the rotation matrix $IT=R_{in}$ is generalized to a non-singular matrix $A_{in}$, i.e. we act an affine transform on the spatial coordinate $X$. Also, we can define special $TA$ for vector fields by limiting $A_{out}=A_{in}$. Due to $RA\subset TA$, in theory, $TA$ can model more complex deformations of multi-channel data. Recently, some papers also generated geometric moment invariants of multi-channel functions to $TA$ and special $TA$ \cite{34,49,50}. However, without the parameters of $IT$, it cannot determine an affine-equivariant region for calculating invariant features around a given point. For example, when detecting vortices in 2D vector fields, the papers\cite{49,50} calculated moment invariants to $TA$ on a circular region with a fixed radius around each point. However, to comply with the definition of $TA$, the circular region should be transformed into an elliptical region using $IT$. This problem will result in calculation errors and impact the performance of moment invariants to $TA$. That's why in this paper we demand $IT$ must be spatial rotation instead of affine transform.    


\subsection{Gaussian-Hermite Moments of Multi-Channel Functions}
\label{section:2.3}
The $p$-order Hermite polynomial is defined as
\begin{equation}\label{equ:4}
	H_{p}(x)=(-1)^{p}\mathrm{exp}(x^{2})\frac{\mathrm{d}^{p}}{\mathrm{d}x^{p}}\mathrm{exp}(-x^{2})
\end{equation}
where $p$ is a non-negative integer and $H_{p}(x)$ satisfies the following recurrence relation 
\begin{equation}\label{equ:5}
	H_{p}(x)=2xH_{p-1}(x)-2(p-1)H_{p-2}(x)
\end{equation}
where $H_{0}(x)=1$ and $H_{1}(x)=2x$. The Hermite polynomials are orthogonal on $(-\infty,+\infty)$ with the weight $\omega(x)=\mathrm{exp}(-x^{2})$
\begin{equation}\label{equ:6}
	\int_{-\infty}^{+\infty}\mathrm{exp}(-x^{2})H_{p_{1}}(x)H_{p_{2}}(x)\mathrm{d}x=p_{1}!2^{p_{1}}\sqrt{\pi}\delta_{p_{1}p_{2}}
\end{equation}
where $p_{1}$ and $p_{2}$ are non-negative integers, and $\delta_{p_{1}p_{2}}$ is the Kronecker delta. 

\begin{figure}
	\centering
	\includegraphics[height=45.5mm,width=82mm]{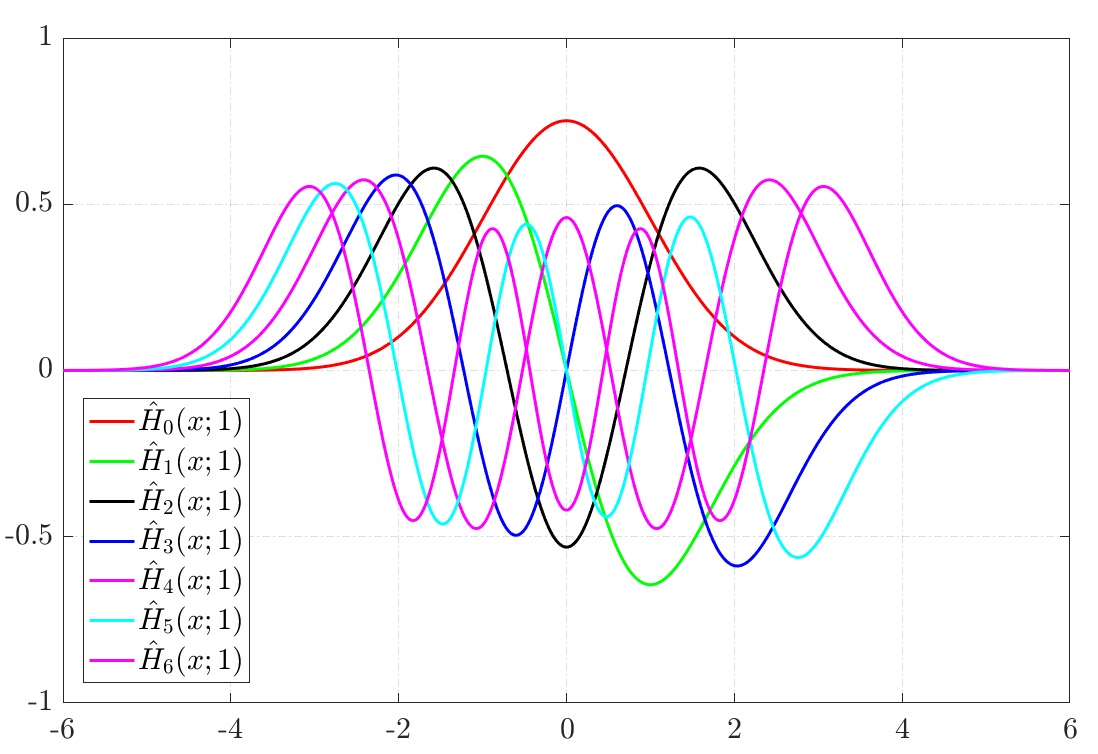}
	\caption{The Gaussian-Hermite polynomials up to the sixth order  $(\sigma=1)$.}
	\label{fig:3}
\end{figure}

Due to a high range of values and poor localization, original Hermite polynomials are difficult to be employed directly without any normalization. To overcome this, some researchers modulated Hermite polynomials with a Gaussian function. In our paper, the $p$-order Gaussian–Hermite polynomial is defined as
\begin{equation}\label{equ:7}
	\begin{split}
		\hat{H}_{p}(x;\sigma)&=\frac{1}{(-\sigma)^{p}}\mathrm{exp}\left(-\frac{x^{2}}{2\sigma^{2}}\right)H_{p}\left(\frac{x}{\sigma}\right)\\&
		=\mathrm{exp}\left(\frac{x^{2}}{2\sigma^{2}}\right)\frac{\mathrm{d}^{p}}{\mathrm{d}x^{p}}\mathrm{exp}\left(-\frac{x^{2}}{\sigma^{2}}\right)\\&
		=G_{2}\cdot \frac{\mathrm{d}^{p}}{\mathrm{d}x^{p}}G_{1}
	\end{split}
\end{equation}
where $G_{1}$ and $G_{2}$ represent two Gaussian functions and $\sigma$ is a user-defined scale parameter which controls the attenuation of the polynomial. When setting $\sigma=1$, we plot Gaussian-Hermite polynomials up to the sixth order in Fig.\ref{fig:3}. Note that in order to show them in a similar range, we normalize $\hat{H}_{p}(x;\sigma)$ by $1/\sqrt{p!2^{p}\sqrt{\pi}}$. 

Given a general multi-channel function $F(X)$ defined by (\ref{equ:1}), we can define its $\left(p_{1}+p_{2}+\cdots +p_{m}\right)$-order Gaussian-Hermite moment as
\begin{equation}\label{equ:8}
	\eta^{n}_{p_{1}p_{2}\cdots p_{M}}=\idotsint \limits_{\Omega}\prod_{m=1}^{M}\hat{H}_{p_{m}}(x_{m};\sigma)f_{n}(X)\mathrm{d}x_{1}\cdots \mathrm{d}x_{M}
\end{equation}
where $p_{1}$, $p_{2},\cdots,p_{M}$ are non-negative integers, and $n\in \{1,2,\cdots,N\}$. The scale parameter $\sigma$ influences the performance of Gaussian-Hermite moments, but there is no exact rule for setting its value. Finding appropriate $\sigma$ is a heuristic which depends on the size and content of the data \cite{46}. 


In \cite{23,27,32,47,48}, Yang et al. defined Gaussian-Hermite moments of grayscale images, 3D shapes, RGB images and 2D vector fields. In fact, all of these moments are special cases of (\ref{equ:8}), because these data can be viewed as different multi-channel functions. 

\section{The structural framework of MGHMIs}
\label{section:3}
Using two fundamental differential operators and two fundamental primitives, this section defines a unified framework to generate $MGHMIs$ of general multi-channel functions to $RA$ and $TR$. 

\subsection{Two fundamental differential operators}
\label{section:3.1}
\newtheorem{definition}{Definition}
\begin{definition}\label{def:1}
Let the symbol $\nabla_{k}$ denote the gradient operator with respect to $X_{k}=(x^{k}_{1},x^{k}_{2},\cdots, x^{k}_{M})^{T}$, which means
\begin{equation}\label{equ:9}
	\nabla_{k}=
	\left(
	\frac{\partial}{\partial{x^{k}_{1}}},
	\frac{\partial}{\partial{x^{k}_{2}}},
	\cdots,
	\frac{\partial}{\partial{x^{k}_{M}}}
	\right)
	^{T}
\end{equation}
For $M$ arbitrary points $X_{1},X_{2},\cdots,X_{M}$, two fundamental differential operators $\phi_{12}$ and $\psi_{12\cdots M}$ are defined as
\begin{equation}\label{equ:10}
	\phi_{12}={\nabla_{1}}^{T}\cdot \nabla_{2}
\end{equation}
and
\begin{equation}\label{equ:11}
\psi_{12\cdots M}=\left|
\left(
\nabla_{1}, \nabla_{2}, \cdots, \nabla_{M}
\right)
\right|
\end{equation}
The symbol $\left|\cdot\right|$ denotes the determinant of a given matrix.
\end{definition}

\newtheorem{lemma}{Lemma}
\begin{lemma}\label{lem:1}
Suppose that $X_{k}$ is transformed into $Y_{k}=(y^{k}_{1},y^{k}_{2},\cdots, y^{k}_{M})^{T}$ using $IT=R_{in}\in\mathbb{R}^{M\times M}$, where $k=1,2,\cdots,M$. Let differential operators $\phi^{'}_{12}$ and $\psi^{'}_{12\cdots M}$ be defined as
\begin{equation}\label{equ:12}
	\begin{split}
		&\phi^{'}_{12}={\nabla^{'}_{1}}^{T}\cdot {\nabla^{'}_{2}}\\&
		\psi^{'}_{12\cdots M}=\left|\left(
			{\nabla^{'}_{1}}, {\nabla^{'}_{2}}, \cdots, {\nabla^{'}_{M}}
		\right)\right|\\
	\end{split}
\end{equation}
where
\begin{equation}\label{equ:13}
	\nabla^{'}_{k}=
	\left(
	\frac{\partial}{\partial{y^{k}_{1}}},
	\frac{\partial}{\partial{y^{k}_{2}}},
	\cdots,
	\frac{\partial}{\partial{y^{k}_{M}}}
	\right)
	^{T}
\end{equation}
We have $\phi^{'}_{12}=\phi_{12}$ and $\psi^{'}_{12\cdots M}=\psi_{12\cdots M}$.
\end{lemma}

\begin{IEEEproof}
According to (\ref{equ:2}), we have $Y_{k}=R_{in} \cdot X_{k}$ and $R_{in}^{-1}=R_{in}^{T}$. Based on the chain's rule of composite functions, the gradient operators $\nabla^{'}_{k}$ and $\nabla_{k}$ are related by
\begin{equation}\label{equ:14}
	\begin{split}
		\nabla^{'}_{k}&=\left(
		\frac{\partial}{\partial{y^{k}_{1}}},
		\frac{\partial}{\partial{y^{k}_{2}}},
		\cdots,
		\frac{\partial}{\partial{y^{k}_{M}}}
		\right)
		^{T}\\&\\&
		=
		\left(
		\begin{array}{cccc}
			\frac{\partial{x^{k}_{1}}}{\partial{y^{k}_{1}}} & \frac{\partial{x^{k}_{2}}}{\partial{y^{k}_{1}}} &
			\cdots &
			\frac{\partial{x^{k}_{M}}}{\partial{y^{k}_{1}}} \vspace{1ex} \\
			\frac{\partial{x^{k}_{1}}}{\partial{y^{k}_{2}}} & \frac{\partial{x^{k}_{2}}}{\partial{y^{k}_{2}}} &
			\cdots &
			\frac{\partial{x^{k}_{M}}}{\partial{y^{k}_{2}}}\\
			\vdots &
			\vdots &
			\ddots &
			\vdots \\
			\frac{\partial{x^{k}_{1}}}{\partial{y^{k}_{M}}} & \frac{\partial{x^{k}_{2}}}{\partial{y^{k}_{M}}} &
			\cdots &
			\frac{\partial{x^{k}_{M}}}{\partial{y^{k}_{M}}}\\
		\end{array}
		\right)
		\left(
		\begin{array}{c}
			\frac{\partial}{\partial{x^{k}_{1}}}  \vspace{1ex} \\
			\frac{\partial}{\partial{x^{k}_{2}}} \\
			\vdots \\
			\frac{\partial}{\partial{x^{k}_{M}}} \\
		\end{array}
		\right)\\&\\&
		=R_{in}\nabla_{k}
	\end{split}
\end{equation}
Substituting the relations $R_{in}^{-1}=R_{in}^{T}$, $\left|R_{in}\right|=1$ and $\nabla^{'}_{k}=R_{in}\nabla_{k}$ into (\ref{equ:12}), we have
\begin{equation}\label{equ:15}
	\begin{split}
		&\phi^{'}_{12}
		=\left(R_{in}\nabla_{1}\right)^{T}R_{in}\nabla_{2}
		={\nabla_{1}}^{T}R^{T}_{in}R_{in}\nabla_{2}
		=\phi_{12}\\&
		\psi^{'}_{12\cdots M}
		=\left|\left(R_{in}\nabla_{1}, \cdots, R_{in}\nabla_{M}\right)\right|\\&~~~~~~~~~~~
		=\left|R_{in}\right|\left|\left(\nabla_{1}, \cdots, \nabla_{M}\right)\right|\\&~~~~~~~~~~~
		=\psi_{12\cdots M}
	\end{split}
\end{equation}
The proof is completed. 
\end{IEEEproof}

Lemma \ref{lem:1} indicates that two fundamental differential operators $\phi_{12}$ and $\psi_{12\cdots M}$ are invariant to $IT=R_{in}$.

\vspace{1ex}
\begin{definition}\label{def:2}
Given a positive integer $K$, we can construct a cumulative product of two fundamental differential operators with respect to $X_{k}=(x^{k}_{1},x^{k}_{2},\cdots x^{k}_{M})^{T}$, where $k=1,2,\cdots,K$. It is also a differential operator and defined as
\begin{equation}\label{equ:16}
D=\prod_{b1,\cdots,b_{M}=1}^{K}(\psi_{b_{1}\cdots b_{M}})^{t_{b_{1}\cdots t_{M}}}\prod^{K}_{a_{1},a_{2}=1}(\phi_{a_{1}a_{2}})^{s_{a_{1}a_{2}}}
\end{equation}   
where
\begin{itemize}
\item  $s_{a_{1}a_{2}}$ and $t_{b_{1}b_{2}\cdots b_{M}}$ are non-negative integers and each $X_{k}$(i.e. $\nabla_{k}$) is involved at least once in $D$.
\item We just consider $a_{1}\leq a_{2}$ because $\phi_{a_{1}a_{2}}=\phi_{a_{2}a_{1}}$. 
\item For the determinant $\psi_{b_{1}b_{2}\cdots b_{M}}$, it is meaningful to consider only $b_{1}<b_{2}<\cdots<b_{M}$. In fact, if all elements of one column are identical with the elements of some other columns, the determinant is zero; and the interchange of any two columns of the determinant only changes its sign.
\end{itemize}
\end{definition}

\subsection{Two fundamental primitives}
\label{section:3.2}
\begin{definition}\label{def:3}
Given a general multi-channel function $F(X)$ defined by (\ref{equ:1}) and $N$ arbitrary points $X_{1},X_{2},\cdots,X_{N}\in\Omega$, two fundamental primitives $\Gamma_{12}$ and $\Lambda_{12\cdots N}$ are defined as
\begin{equation}\label{equ:17}
\Gamma_{12}=F(X_{1})^{T}\cdot F(X_{2})
\end{equation}
and
\begin{equation}\label{equ:18}
\Lambda_{12\cdots N}=\left|\left(F(X_{1}),F(X_{2}),\cdots, F(X_{N})\right)\right|
\end{equation}
\end{definition}

\begin{lemma}\label{lem:3}
Let a multi-channel function $F(X)$ be transformed into $G(Y)$ using $RA$ defined by (\ref{equ:2}). Suppose that $Y_{k}=(y^{k}_{1},y^{k}_{2},\cdots,y^{k}_{M})^{T}\in\Omega^{'}$ is the corresponding point of $X_{k}\in \Omega$, where $k=1,2,\cdots,N$. We have   
\begin{equation}\label{equ:19}
\Lambda^{'}_{12\cdots N}=\left|A_{out}\right|\Lambda_{12\cdots N}
\end{equation}
where
\begin{equation}\label{equ:20}
\Lambda^{'}_{12\cdots N}=\left|\left(G(Y_{1}),G(Y_{2}),\cdots,G(Y_{N})\right)\right|
\end{equation}
If the non-singular matrix $A_{out}$ in $OT$ degenerates into a rotation matrix $R_{out}\in\mathbb{R}^{N\times N}$, we further have
\begin{equation}\label{equ:21}
\Gamma^{'}_{12}=\Gamma_{12},~~~~
\Lambda^{'}_{12\cdots N}=\Lambda_{12\cdots N}
\end{equation}
where 
\begin{equation}\label{equ:22}
\Gamma^{'}_{12}=G(Y_{1})^{T}\cdot G(Y_{2})
\end{equation}
\end{lemma}

\begin{IEEEproof}
According to (\ref{equ:2}), we have $G(Y_{k})=A_{out} \cdot F(X_{k})$, where $k=1,2,\cdots,N$. As stated in Section \ref{section:2.2}, we omit the outer translation $T_{out}$. Thus, 
\begin{equation}\label{equ:23}
\begin{split}
\Lambda^{'}_{12\cdots N}&=\left|\left(A_{out} \cdot F(X_{1}),\cdots, A_{out} \cdot F(X_{N})\right)\right|\\&=\left|A_{out}\right|\left|\left(F(X_{1}),\cdots, F(X_{N})\right)\right|\\&=\left|A_{out}\right|\Lambda_{12\cdots N}
\end{split}
\end{equation} 
When $A_{out}$ degenerates into $R_{out}$, we have $R_{out}^{-1}=R_{out}^{T}$ and $\left|R_{out}\right|=1$ Substituting these properties into (\ref{equ:22}) and (\ref{equ:23}), we further get
\begin{equation}\label{equ:24}
\begin{split}
&\Gamma^{'}_{12}=F(X_{1})^{T}R^{T}_{out}R_{out}F(X_{2})=\Gamma_{12}\\&
\Lambda^{'}_{12\cdots N}=\left|R_{out}\right|\Lambda_{12\cdots N}=\Lambda_{12\cdots N}
\end{split}
\end{equation} 
The proof is completed.
\end{IEEEproof}

Lemma \ref{lem:3} demonstrates that the fundamental primitive $\Lambda_{12\cdots N}$ is relatively invariant to $RA$. When the non-singular matrix $A_{out}$ in $OT$ degenerates into a rotation matrix $R_{out}$, both $\Gamma_{12}$ and $\Lambda_{12\cdots N}$ are absolutely invariant to $TR$. Some previous papers used specific instances of these two fundamental primitives to generate geometric moment invariants of different types of multi-channel data \cite{26,34,49,50}. These works inspire us. 
\vspace{1ex}
\begin{definition}\label{def:4}
Given a general multi-channel function $F(X)$ defined by (\ref{equ:1}) and an integer $K$, we can construct the cumulative product of two fundamental primitives with respect to $K$ arbitrary points $X_{1},X_{2},\cdots,X_{K}\in \Omega$
\begin{equation}\label{equ:25}
P=\prod_{d_{1},\cdots,d_{N}=1}^{K}\left(\Lambda_{d_{1}\cdots d_{N}}\right)^{t_{d_{1}\cdots d_{N}}}\prod_{c_{1},c_{2}=1}^{K}\left(\Gamma_{c_{1}c_{2}}\right)^{s_{c_{1}c_{2}}}
\end{equation} 
where
\begin{itemize}
\item $s_{c_{1}c_{2}}$ and $t_{d_{1}d_{2}\cdots d_{N}}$ are non-negative integers. Each $F(X_{k})$ is used once and only once in $P$. This implies that $s_{c_{1}c_{2}}$ and $t_{d_{1}d_{2}\cdots d_{N}}$ can only equal 0 or 1. 
\item Similar to (\ref{equ:16}), we only consider $c_{1}<c_{2}$ and $d_{1}<d_{2}<\cdots<d_{N}$. Here, we demand $c_{1}\neq c_{2}$ because $F(X_{k})$ can only be used once in $P$.   
\end{itemize}
\end{definition}
	
\subsection{The construction of MGHMIs}
\label{section:3.3}
Let us now design a structural framework of $MGHMIs$ based on definitions and lemmas in Section \ref{section:3.1} and \ref{section:3.2}.  

\begin{definition}\label{def:5}
Given a general multi-channel function $F(X)$ defined by (\ref{equ:1}), suppose that $K$ arbitrary points $X_{1},X_{2},\cdots,X_{K}\in \Omega$. Then, we can define $MGHMI$ as
\begin{equation}\label{equ:26}
MGHMI=\idotsint \limits_{\Omega^{K}}G_{2}\cdot D(G_{1})\cdot P\prod_{k=1}^{K}\prod_{m=1}^{M}\mathrm{d}x^{k}_{m}
\end{equation}
where $D$ and $P$ are given by (\ref{equ:16}) and (\ref{equ:25}); two Gaussian functions $G_{1}$ and $G_{2}$ are defined as 
\begin{equation}\label{equ:27}
G_{1}=\prod_{k=1}^{K}\mathrm{exp}\left(-\frac{X^{T}_{k}X_{k}}{\sigma^{2}}\right),~~
G_{2}=\prod_{k=1}^{K}\mathrm{exp}\left(\frac{X^{T}_{k}X_{k}}{2\sigma^{2}}\right)
\end{equation}
and the scale parameter $\sigma$ is a positive real number. Note that $D(G_{1})$ means to act the differential operator $D$ on the function $G_{1}$.  
\end{definition}

\newtheorem{theorem}{Theorem}
\begin{theorem}\label{the:1}
Let $F(X)$ be transformed into $G(Y)$ using $RA$ defined by (\ref{equ:2}) and $Y_{k}\in \Omega^{'}$ be the corresponding point of $X_{k}\in\Omega$, where $k=1,2,...,K$. Then, the following two relations can be derived. 
\begin{itemize}
\item When all $s_{c_{1}c_{2}}$ in $P$ are equal to $0$, we have   
\begin{equation}\label{equ:28}
MGHMI^{'}=\left|A_{out}\right|^{\sum t_{d_{1}d_{2}\cdots d_{N}}}MGHMI
\end{equation}
\item When $RA$ degenerates into $TR$, which means that the non-singular matrix $A_{out}$ in $OT$ degenerates into a rotation matrix $R_{out}$, we have
\begin{equation}\label{equ:29}
MGHMI^{'}=MGHMI
\end{equation}
\end{itemize}

Note that $MGHMI$ is defined by (\ref{equ:26}) and $MGHMI^{'}$ is defined in a similar fashion
\begin{equation}\label{equ:30}
MGHMI^{'}=\idotsint \limits_{{\Omega^{'}}^{K}}G^{'}_{2}\cdot D^{'}\left(G^{'}_{1}\right)\cdot P^{'}\prod_{k=1}^{K}\prod_{m=1}^{M}\mathrm{d}y^{k}_{m}
\end{equation} 
where
\begin{equation}\label{equ:31}
D^{'}=\prod_{b1,\cdots,b_{M}=1}^{K}(\psi^{'}_{b_{1}\cdots b_{M}})^{t_{b_{1}\cdots b_{M}}}\prod^{K}_{a_{1},a_{2}=1}(\phi^{'}_{a_{1}a_{2}})^{s_{a_{1}a_{2}}}
\end{equation}
\begin{equation}\label{equ:32}
P^{'}=\prod_{d_{1},\cdots,d_{N}=1}^{K}\left(\Lambda^{'}_{d_{1}\cdots d_{N}}\right)^{t_{d_{1}\cdots d_{N}}}\prod_{c_{1},c_{2}=1}^{K}\left(\Gamma^{'}_{c_{1}c_{2}}\right)^{s_{c_{1}c_{2}}}
\end{equation}
\begin{equation}\label{equ:33}
G^{'}_{1}=\prod_{k=1}^{K}\mathrm{exp}\left(-\frac{Y^{T}_{k}Y_{k}}{\sigma^{2}}\right),~~
G^{'}_{2}=\prod_{k=1}^{K}\mathrm{exp}\left(\frac{Y^{T}_{k}Y_{k}}{2\sigma^{2}}\right)
\end{equation}
and $\phi^{'}_{a_{1}a_{2}}$, $\psi^{'}_{b_{1}b_{2}\cdots b_{M}}$, $\Gamma^{'}_{c_{1}c_{2}}$ and $\Lambda^{'}_{d_{1}d_{2}\cdots d_{N}}$ are defined by (\ref{equ:12}), (\ref{equ:20}) and (\ref{equ:22}), respectively.
\end{theorem}

\begin{IEEEproof}
First, by substituting (\ref{equ:15}) into (\ref{equ:31}), we can get $D^{'}=D$. Meanwhile, since $Y_{k}=R_{in} \cdot X_{k}$ and $R^{-1}_{in}=R^{T}_{in}$, we have 
\begin{equation}\label{equ:34}
	Y^{T}_{k}Y_{k}=X^{T}_{k}R^{T}_{in}R_{in}X_{k}=X^{T}_{k}R^{-1}_{in}R_{in}X_{k}=X^{T}_{k}X_{k}
\end{equation} 
meaning that $G^{'}_{1}=G_{1}$ and $G^{'}_{2}=G_{2}$. Hence, the following relation can be obtained
\begin{equation}\label{equ:35}
G^{'}_{2}\cdot D^{'}(G^{'}_{1})=G_{2}\cdot D(G_{1})
\end{equation} 

Then, using the property $\left|R_{in}\right|=1$, we get
\begin{equation}\label{equ:36}
	\begin{split}
		\prod_{k=1}^{K}\prod_{m=1}^{M}\mathrm{d}y^{k}_{m}&=\prod_{k=1}^{K}abs\left(\left|J_{k}\right|\right)\prod_{m=1}^{M}\mathrm{d}x^{k}_{m}\\&=abs\left(\left|R_{in}\right|\right)^{K}\prod_{k=1}^{K}\prod_{m=1}^{M}\mathrm{d}x^{k}_{m}\\&=\prod_{k=1}^{K}\prod_{m=1}^{M}\mathrm{d}x^{k}_{m}
	\end{split}
\end{equation} 
where the symbol $J_{k}$ denotes the Jacobian matrix with respect to $X_{k}$ and $Y_{k}$, and the function $abs(\cdot)$ returns the absolute value of a given number.
\\
\\
\noindent
\textbf{(1):} According to (\ref{equ:23}), when all $s_{c_{1}c_{2}}$ are equal to $0$, $P^{'}$ defined by (\ref{equ:32}) becomes 
\begin{equation}\label{equ:37}
	\begin{split}
		P^{'}&=\prod_{d_{1},\cdots,d_{N}=1}^{K}\left(\Lambda^{'}_{d_{1}\cdots d_{N}}\right)^{t_{d_{1}\cdots d_{N}}}\\&=\prod_{d_{1},\cdots,d_{N}=1}^{K}\left|A_{out}\right|^{t_{d_{1}\cdots d_{N}}}\left(\Lambda_{d_{1}\cdots d_{N}}\right)^{t_{d_{1}\cdots d_{N}}}\\&
		=\left|A_{out}\right|^{\sum t_{d_{1}\cdots d_{N}}}\prod_{d_{1},\cdots,d_{N}=1}^{K}\left(\Lambda_{d_{1}\cdots d_{N}}\right)^{t_{d_{1}\cdots d_{N}}}\\&
		=\left|A_{out}\right|^{\sum t_{d_{1}\cdots d_{N}}}P
	\end{split}
\end{equation} 
Substituting (\ref{equ:35}), (\ref{equ:36}) and (\ref{equ:37}) into (\ref{equ:30}), we have 
\begin{equation}\label{equ:38}
MGHMI^{'}=\left|A_{out}\right|^{\sum t_{d_{1}\cdots d_{N}}}MGHMI
\end{equation}
\\
\noindent
\textbf{(2):} Using (\ref{equ:24}), when $RA$ degenerates into $TR$, we can get
\begin{equation}\label{equ:39}
\begin{split}
P^{'}&=\prod_{d_{1},\cdots,d_{N}=1}^{K}\left(\Lambda^{'}_{d_{1}\cdots d_{N}}\right)^{t_{d_{1}\cdots d_{N}}}\prod_{c_{1},c_{2}=1}^{K}\left(\Gamma^{'}_{c_{1}c_{2}}\right)^{s_{c_{1}c_{2}}}\\&
=\prod_{d_{1},\cdots,d_{N}=1}^{K}\left(\Lambda_{d_{1}\cdots d_{N}}\right)^{t_{d_{1}\cdots d_{N}}}\prod_{c_{1},c_{2}=1}^{K}\left(\Gamma_{c_{1}c_{2}}\right)^{s_{c_{1}c_{2}}}\\&
=P
\end{split}
\end{equation}
Substituting (\ref{equ:35}), (\ref{equ:36}) and (\ref{equ:39}) into (\ref{equ:30}), we have 
\begin{equation}\label{equ:40}
MGHMI^{'}=MGHMI
\end{equation}
The proof is completed.
\end{IEEEproof}

Theorem \ref{the:1} proves that $MGHMIs$ defined by (\ref{equ:26}) are absolutely invariant to $TR$, and relatively invariant to $RA$ by setting some parameters to $0$. 

Moreover, an $MGHMI$ can be expressed as a $K$-degree polynomial in Gaussian-Hermeite moments of $F(X)$. In fact, the expansion of $D$ is a polynomial of $\partial^{p}/\partial{(x^{k}_{m})^{p}}$. According to (\ref{equ:7}), we find that the expansion of $G_{2}\cdot D(G_{1})$ can be directly obtained by replacing $\partial^{p}/\partial{(x^{k}_{m})^{p}}$ in the expansion of $D$ with $\hat{H}_{p}(x^{k}_{m};\sigma)$. Then, by substituting the expansions of $G_{2}\cdot D(G_{1})$ and $P$ into (\ref{equ:26}), we can expand $MGHMI$ as a polynomial of $\eta^{n}_{p_{1}p_{2}\cdots p_{M}}$ defined by (\ref{equ:8}). 

To better explain this, we generate an $MGHMI$ of 2D vector fields $(M=N=2)$ by setting $D=\psi_{12}$ and $P=\Lambda_{12}$. First, the differential operator $D=\psi_{12}$ can be expanded as  
\begin{equation}\label{equ:41}
D=\left|\left(\nabla_{1}, \nabla_{2}\right)\right|=\frac{\partial^{2}}{\partial{x^{1}_{1}}\partial{x^{2}_{2}}}-\frac{\partial^{2}}{\partial{x^{1}_{2}}\partial{x^{2}_{1}}}
\end{equation}
By replacing $\partial^{p}/\partial{(x^{k}_{m})^{p}}$ with $\hat{H}_{p}(x^{k}_{m};\sigma)$, we have  
\begin{equation}\label{equ:42}
G_{2}\cdot D(G_{1})=\hat{H}_{1}(x^{1}_{1};\sigma)\hat{H}_{1}(x^{2}_{2};\sigma)-\hat{H}_{1}(x^{1}_{2};\sigma)\hat{H}_{1}(x^{2}_{1};\sigma)
\end{equation}
Further, $P=\Lambda_{12}$ can be expanded as 
\begin{equation}\label{equ:43}
P=\left|F(X_{1}), F(X_{2})\right|=f_{1}(X_{1})f_{2}(X_{2})-f_{2}(X_{1})f_{1}(X_{2})
\end{equation}
Substituting (\ref{equ:42}) and (\ref{equ:43}) into (\ref{equ:26}), we finally have
\begin{equation}\label{equ:44}
\begin{split}
MGHMI&=\idotsint \limits_{\Omega^{2}}G_{2}\cdot D(G_{1})\cdot P~\mathrm{d}x^{1}_{1}\mathrm{d}x^{1}_{2}\mathrm{d}x^{2}_{1}\mathrm{d}x^{2}_{2}\\&
=2\cdot\left(\eta^{1}_{10}\eta^{2}_{01}-\eta^{1}_{01}\eta^{2}_{10}\right)
\end{split}
\end{equation}

The order of $MGHMI$ is denoted as $O$, which is the highest order of $\eta^{n}_{p_{1}p_{2}\cdots p_{M}}$ that $MGHMI$ depends upon. This value indicates that in the differential operator $D$ defined by (\ref{equ:16}), at least one $k$ appears $O$ times in all subscripts $a_{1}a_{2}$ and $b_{1}b_{2}\cdots b_{M}$, where $k=1,2,\cdots,K$. For the above $MGHMI$, we have the degree $K=2$ and the order $O=1$.  

In addition, Theorem \ref{the:1} illustrates that $MGHMIs$ are just relatively invariant to $RA$. To eliminate $\left|A_{out}\right|^{\sum t_{d_{1}\cdots d_{N}}}$ and obtain an absolute invariant, we can normalize a relative invariant by other relative invariants or by proper sum/power of them so that Jacobians get canceled. In fact, this kind of normalization approach has be used in many previous papers \cite{34,49,50,53}. 


\section{The instances of MGHMIs}
\label{section:4}
For a specific type of multi-channel data, we can use Definition \ref{def:5} to generate all possible $MGHMIs$ up to the degree $K$ and the order $O$. The generation procedure can be described as following four steps. Note that some symbolic computation software can help us to achieve each of them.  
\begin{itemize}
	\item \textbf{Step 1}: Since the degree of each $MGHMI$ is less than or equal to $K$, at most $K$ points can be used to construct the differential operator $D$ in Definition \ref{def:2} and $P$ in Definition \ref{def:4}. We generate all possible fundamental differential operators $\phi_{a_{1}a_{2}}$ and $\psi_{b_{1}b_{2}\cdots b_{M}}$ with respect to $X_{1},X_{2},..., X_{K}$, and the set of them is denoted as $U$. Similarly, we derive the set $V$ of all possible fundamental primitives $\Gamma_{c_{1}c_{2}}$ and $\Lambda_{d_{1}d_{2}\cdots d_{N}}$. Taking 2D vector fields as an example, when setting $K=2$, we have $U=\left\{\phi_{11},\phi_{12},\phi_{22},\psi_{12}\right\}$ and $V=\left\{\Gamma_{12},\Lambda_{12}\right\}$. Note that when generating $MGHMIs$ invariant to $RA$, we need to remove all $\Gamma_{c_{1}c_{2}}$ from $V$. 
	\item \textbf{Step 2}: Based on the sets $U$ and $V$, we can further generate $\widetilde{U}=\bigcup\limits_{r=1}^{R_{1}}U^{r}$ and $\widetilde{V}=\bigcup\limits_{r=1}^{R_{2}}V^{r}$, where $U^{r}$ $\left(V^{r}\right)$ represents the cumulative Cartesian product of $U$ $\left(V\right)$. Each of the elements in $U^{r}$ $\left(V^{r}\right)$ contains $r$ fundamental differential operators (fundamental primitives), which are used for constructing $D$ ($P$). We have two constrains to determine the values of $R_{1}$ and $R_{2}$, respectively. \textbf{(1)} The order of each $MGHMI$ must be less than or equal to $O$. Thus, for each of elements in the set $U^{R_{1}+1}$, we demand that at least one $k$ appears more than $O$ times in all subscripts $a_{1}a_{2}$ and $b_{1}b_{2}\cdots b_{M}$, where $k\in\{1,2,...,K\}$. \textbf{(2)} As stated in Definition \ref{def:4}, each of $F(X_{k})$ must be used only once in $P$. Therefore, we also demand that all elements in $V^{R_{2}+1}$ do not satisfy this condition.     
	\item \textbf{Step 3}: Obviously, each of elements in $\widetilde{U}\times\widetilde{V}$ represents a pair of $D$ and $P$ used in (\ref{equ:26}). Since $D$ and the corresponding $P$ must be constructed using the same points $X_{k}$, those elements that do not meet this requirement can also be discarded.
	\item \textbf{Step 4}: Using the procedure described in Section \ref{section:3.3}, we expanded $MGHMIs$ as polynomials in terms of Gaussian-Hermite moments. The expansions of some $MGHMIs$ are $0$, and one $MGHMI$ can also be generated repeatedly. We further eliminate these meaningless invariants.     
\end{itemize}
However, the above procedure does not guarantee that there are no dependent invariants in a generated set, which means some of $MGHMIs$ are the functions of the others. In many practical applications, a single real-valued $MGHMI$ does not provide enough discriminative ability, and all $MGHMIs$ in the set must be used simultaneously as a feature vector. Previous research has found that dependent moment invariants contribute nothing to the discrimination power of the vector but also increase the dimensionality of feature space and computational time \cite{46}. Thus, identifying and discarding dependent $MGHMIs$ in the set is highly desirable. In \cite{54}, we have introduced some approaches to eliminate linear, polynomial and functional dependencies among elements in a set. They can also be used to handle the set of $MGHMIs$. 

\begin{table}
		\caption{\label{table:2} The structural information of independent $MGHMI$ sets of various multi-channel data. "Number" means the number of $MGHMIs$ in an independent set.}
		\centering
		\begin{tabular}{p{2.5cm}p{1.4cm}p{0.8cm}p{0.8cm}p{1.2cm}}
				\toprule[1.1pt]
				\textbf{Data Type}&\textbf{Transform}&\textbf{K}&\textbf{O}&\textbf{Number}\\
				\toprule[1.1pt]
				RGB Images&$RA$&$3$&$3$&13\\
				2D Vector Fields&$TR$&$2$&$3$&7\\
				2D Vector Fields&$RA$&$2$&$3$&7\\
				Color Volume Data&$RA$&$3$&$3$&14\\
				\bottomrule
			\end{tabular}
\end{table}

\begin{table}
		\caption{\label{table:3} Seven $MGHMIs$ of 2D vector fields, which are invariant to $TR$. They form an independent set.}
		\centering
		\begin{tabular}{p{0.4cm}p{7.5cm}}
				\toprule[1.1pt]
				\textbf{No.}&\textbf{The Expansion}\\
				\toprule[1.1pt]
				\bm{$1$}&$\left(\eta^{1}_{10}\right)^{2}+\left(\eta^{1}_{01}\right)^{2}+\left(\eta^{2}_{10}\right)^{2}+\left(\eta^{2}_{01}\right)^{2}$\\
				\midrule
				\bm{$2$}&\makecell[l]{$\eta^{1}_{10}\eta^{1}_{30}+\eta^{1}_{10}\eta^{1}_{12}+\eta^{1}_{01}\eta^{1}_{21}+\eta^{1}_{01}\eta^{1}_{03}+\eta^{2}_{10}\eta^{2}_{30}+\eta^{2}_{10}\eta^{2}_{12}$\\$+\eta^{2}_{01}\eta^{2}_{21}+\eta^{2}_{01}\eta^{2}_{03}$}\\
				\midrule
				\bm{$3$}&$\left(\eta^{1}_{20}\right)^{2}+2\left(\eta^{1}_{11}\right)^{2}+\left(\eta^{1}_{02}\right)^{2}+\left(\eta^{2}_{20}\right)^{2}+2\left(\eta^{2}_{11}\right)^{2}+\left(\eta^{2}_{02}\right)^{2}$\\
				\midrule
				\bm{$4$}&\makecell[l]{$\eta^{1}_{01}\eta^{1}_{12}+\eta^{1}_{01}\eta^{1}_{30}-\eta^{1}_{10}\eta^{1}_{03}-\eta^{1}_{10}\eta^{1}_{21}+\eta^{2}_{01}\eta^{2}_{12}+\eta^{2}_{01}\eta^{2}_{30}$\\$-\eta^{2}_{10}\eta^{2}_{03}-\eta^{2}_{10}\eta^{2}_{21}$}\\
				\midrule
				\bm{$5$}&$\eta^{1}_{20}\eta^{1}_{02}-\left(\eta^{1}_{11}\right)^{2}+\eta^{2}_{20}\eta^{2}_{02}-\left(\eta^{2}_{11}\right)^{2}$\\
				\midrule
				\bm{$6$}&\makecell[l]{$\left(\eta^{1}_{30}\right)^{2}+3\left(\eta^{1}_{21}\right)^{2}+3\left(\eta^{1}_{12}\right)^{2}+\left(\eta^{1}_{03}\right)^{2}+\left(\eta^{2}_{30}\right)^{2}+3\left(\eta^{2}_{21}\right)^{2}$\\$+3\left(\eta^{2}_{12}\right)^{2}+\left(\eta^{2}_{03}\right)^{2}$}\\
				\midrule
				\bm{$7$}&\makecell[l]{$\eta^{1}_{03}\eta^{1}_{21}-\left(\eta^{1}_{12}\right)^{2}+\eta^{1}_{12}\eta^{1}_{30}-\left(\eta^{1}_{21}\right)^{2}+\eta^{2}_{03}\eta^{2}_{21}-\left(\eta^{2}_{12}\right)^{2}$\\$+\eta^{2}_{12}\eta^{1}_{30}-\left(\eta^{2}_{21}\right)^{2}$}\\
				\midrule
		\end{tabular}
\end{table}

For three types of widely used multi-channel data, RGB images, 2D vector fields and color volume data, we derive their functionally independent sets of $MGHMIs$ up to the degree $K$ and the order $O$. The structural information is listed in Table \ref{table:2}. As stated in Section \ref{section:2.2}, $RA$ can model spatial rotation and color variation of RGB images and color volume data. Hence, we generate $MGHMIs$ of these data to $RA$. From a practical point of view, Section \ref{section:2.2} has explained why constructing $MGHMIs$ of vector fields to $TR$ and $RA$ is meaningful. To better verify this, we will conduct vortex detection experiments on real 2D vector fields in Section \ref{section:5.2} and compare the performance of Gaussian-Hermite moment invariants to special $TR$, $TR$ and $RA$. Thus, we generate $MGHMIs$ of 2D vector fields to $TR$ and $RA$, respectively. It is worth noting that a generated set of $MGHMIs$ invariant to $TR$ contains some $MGHMIs$ invariant to $RA$. We first remove these invariants and then eliminate dependencies among the remaining ones. This ensures that two independent sets to $TR$ and $RA$ do not intersect. As an instance, Table \ref{table:3} shows the explicit expansions of seven $MGHMIs$ of 2D vector fields to $TR$, and these invariants form an independent set. Note that the expansions of $MGHMIs$ in the other three independent sets can be found in our Supplementary Materials.   

\section{Experiments and discussion}
\label{section:5}
In this section, we conduct experiments on two types of multi-channel data, RGB images and 2D vector fields, to verify the stability and discriminability of $MGHMIs$ generated in Section \ref{section:4}. Further, we test their robustness to additive noise and their recognition ability in RGB image classification and vortex detection. Some existing moment invariants of multi-channel data are chosen for comparison.

\subsection{MGHMIs of RGB images to RA}
\label{section:5.1}
As shown in Table \ref{table:2}, for RGB images, we derive thirteen $MGHMIs$ up to the third degree and the third order, which form an independent set. According to Theorem \ref{the:1}, for any invariants in this set, we have $MGHMI^{'}=\left|A_{out}\right|MGHMI$ because they are all constructed using the same $P=\Lambda_{123}$. The sum of these invariants also holds this property. Hence, we can normalize each $MGHMI$ by the sum of all invariants in the set. This normalization operation eliminates the constant $\left|A_{out}\right|$ and makes $MGHMIs$ absolutely invariant to $RA$. 

\begin{figure}
	\centering
	\subfloat[Ten original images from the USC-SIPI image database.]
	{\includegraphics[height=33mm,width=82mm]{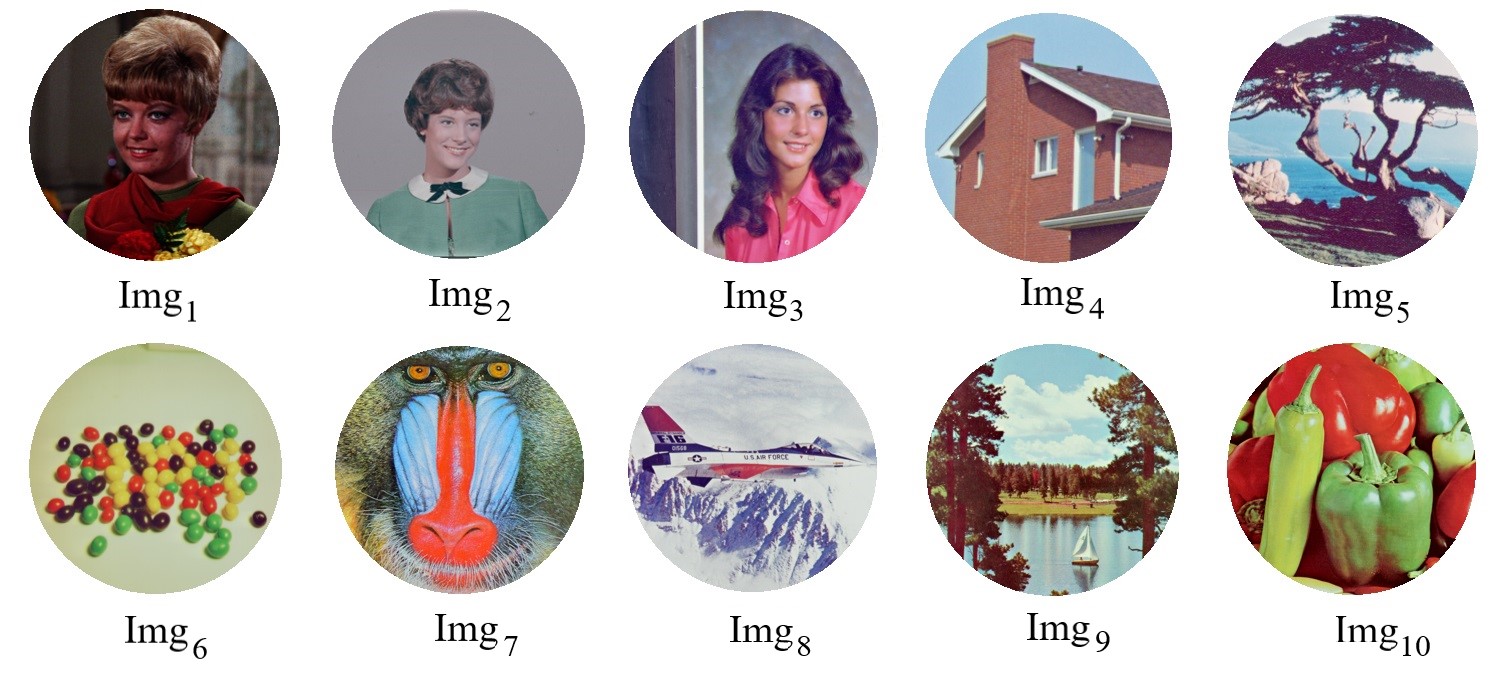}\label{figure:4(a)}\hfill}\\
	\subfloat[Some test images disturbed by different noise levels.]
	{\includegraphics[height=35mm,width=75mm]{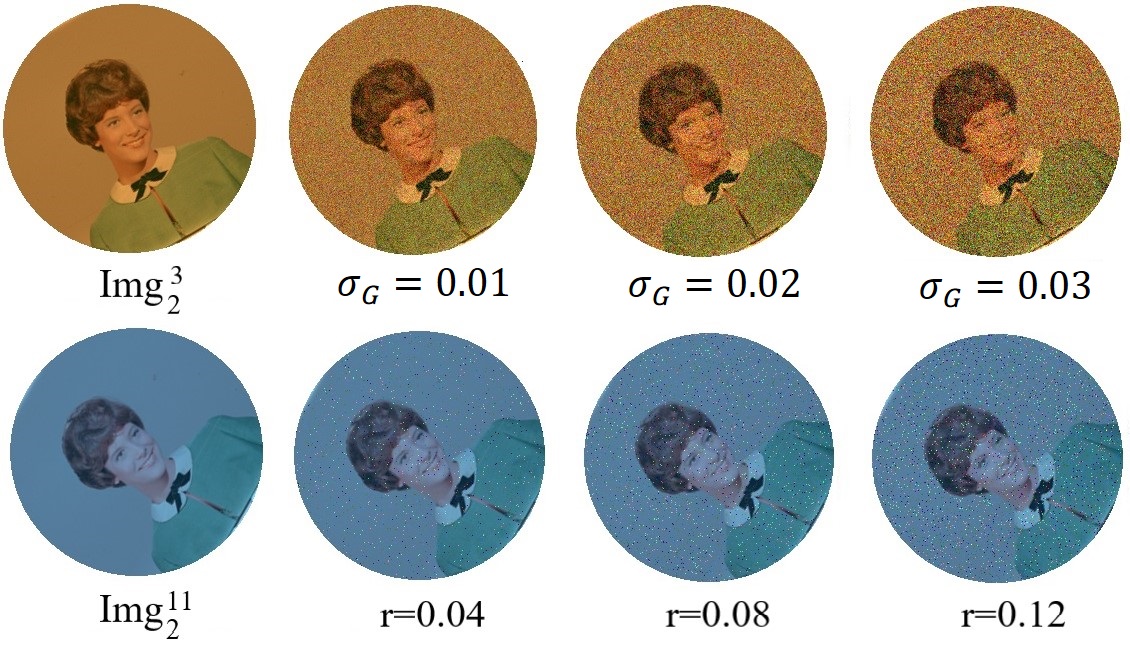}\label{figure:4(b)}\hfill}
	\caption{Synthetic RGB images used for classification.}\label{figure:4}
\end{figure}

In this subsection, we test the performance of these thirteen invariants in RGB image classification. Ten color images are randomly selected from the USC-SIPI image dataset (\url{https://sipi.usc.edu/database/}), which are denoted as Img$_{i}$, where $i=1,2,...,10$. They are used as training images, and each is scaled to $257\times 257$ pixels. Also, for each image, we change the origin of the image coordinate system to the center $(129,129)$, and only the content located in the area of an inscribed circle $(x-129)^{2}+(y-129)^{2}\leq 128$ is used for feature extraction (see Fig.\ref{figure:4(a)}). 

As stated in Section \ref{section:2.2}, $RA$ can simulate realistic deformations of color images caused by viewpoint and illumination changes. Hence, using the formula (\ref{equ:3}), we synthetically generate sixty $RA$ versions Img$^{j}_{i}$ of Img$_{i}$, where $j=1,2,...,60$. Specifically, each training image is first rotated around its center by angles from $0$ to $2\pi$ every $2\pi/60$, and sixty rotated versions are further transformed using sixty color affine transformations. These 3D affine transformations of color space, i.e. $OT$, are randomly generated. To ensure that $(R(x,y),G(x,y),B(x,y))$ is not mapped outside of RGB cube $[0,1]^{3}$ as far as possible, we add some constraints to parameters of $OT$. In fact, according to QR decomposition, a $3\times3$ matrix $A_{out}$ can be expressed as $R_{x} \cdot R_{y} \cdot R_{z} \cdot U$, where the first three ones represent rotation matrices about the x-axis, y-axis and z-axis, respectively, and $U$ is defined as  
\begin{equation}\label{equ:45}
\begin{split}
U=\left(
\begin{array}{ccc}
	S_{x} & M_{1} & M_{2}\\
	0 & S_{y} & M_{3}\\
	0 & 0 & S_{z}\\
\end{array}
\right)
\end{split}
\end{equation}
When generating these matrices, we damend that $\theta_{x},\theta_{y},\theta_{z}\in [-\pi/10,\pi/10]$, which are the angles of $R_{x}$, $R_{y}$ and $R_{z}$, respectively; $S_{x}, S_{y}, S_{z} \in [0.7, 0.9]$; $M_{1},M_{2},M_{3}\in [-0.1,0.1]$. For the outer translation $T_{out}$, we have $t_{1},t_{2},t_{3} \in [-0.1,0.1]$. Note that some $RA$ versions of Img$_{2}$ have been shown in Fig.\ref{figure:2(a)}.   

\begin{figure}
	\centering
	\includegraphics[height=45mm,width=65mm]{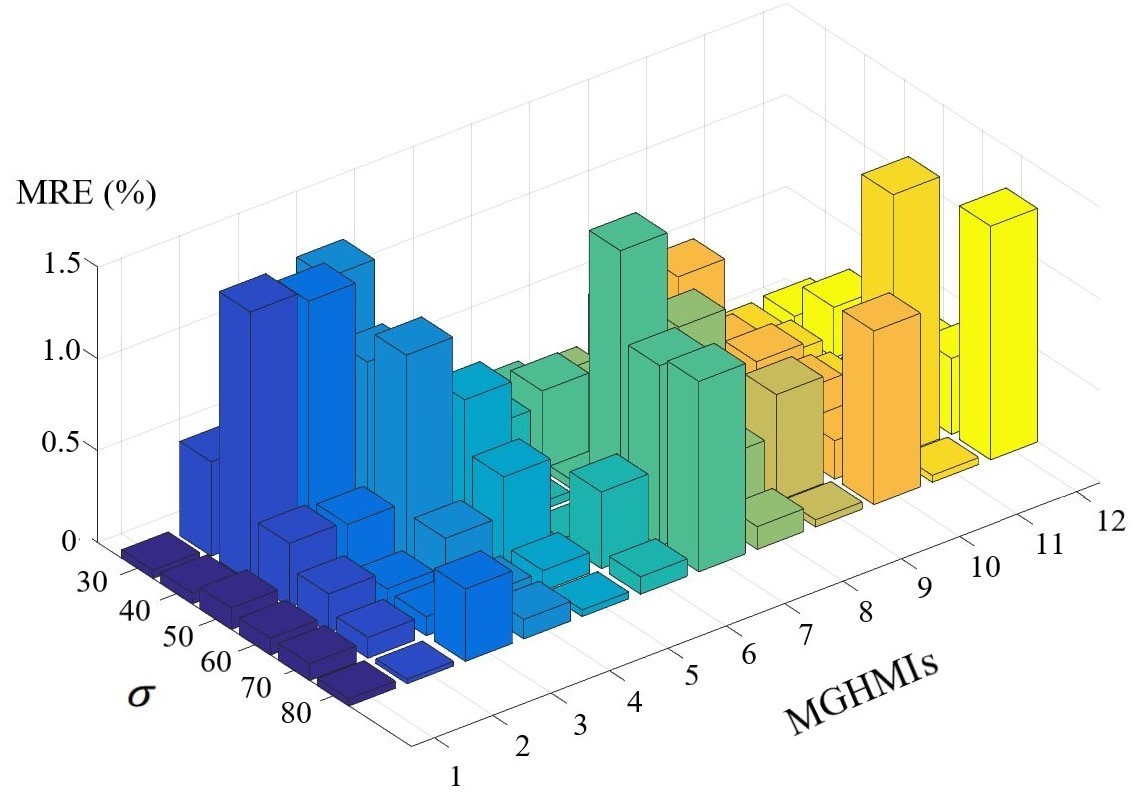}\\
	\caption{The $MRE$ of $MGHMIs$ of RGB images.}\label{figure:5}
\end{figure}
   
The above process yields $10\times 60=600$ test images, and then the numerical values of thirteen $MGHMIs$ are calculated from each image (including training images and test ones). Before that, we subtract the per-channel mean from the image, eliminating three outer translation parameters. As stated in Section \ref{section:2.3}, the scale parameter $\sigma$ affects the performance of Gaussian-Hermite moments defined by (\ref{equ:8}). To find an appropriate value, we set $\sigma=30,40,...,80$. When setting $\sigma<30 \approx 257/9$, $\hat{H}_{p}(x;\sigma)$ defined by (\ref{equ:7}) decays too rapidly on the domain $[-128,128]$ and cannot cover the whole image, which leads to information loss. If $\sigma>80 \approx 257/3$, $\hat{H}_{p}(x;\sigma)$ decays too slowly and some non-zero parts are outside $[-128,128]$, which also decreases the ability of Gaussian-Hermite moments. As stated above, the values of thirteen $MGHMIs$ are normalized by their sum. These normalized $MGHMIs$ are linearly dependent because their sum equals $1$. Thus, we discard the last one and use the first twelve ones as the final feature vector.  

First, the mean relative error $(MRE)$ is used to evaluate the numerical stability of each normalized $MGHMI$, which defined by
\begin{equation}\label{equ:41}
\begin{split}
&MRE=\left(\frac{1}{600}\sum^{10}_{i=1}\sum_{j=1}^{60}E_{ij}\right)\times 100\%\\&
E_{ij}=\frac{\left|MGHMI(\mbox{Img}^{j}_{i})-MGHMI(\mbox{Img}_{i})\right|}{\left|MGHMI(\mbox{Img}_{i})\right|}
\end{split}
\end{equation}
In Fig.\ref{figure:5}, we visualize the $MRE$ of twelve normalized $MGHMIs$ with different $\sigma$. It can be observed that most of them are less than $1\%$, and even the maximum value is only $1.47\%$. These small errors are caused by image resampling. This result illustrates that $MGHMIs$ of RGB images generated by our method are invariant to $RA$. 

Then, to verify the discriminability of these normalized $MGHMIs$, we utilized the Nearest Neighbor classifier for image classification. The Chi-Square distance was used to measure the similarity of two images in the space of features. In fact, the magnitudes of different moment invariants are not the same. As a result, the Euclidean distance between two feature vectors will be dominated by certain invariants with large magnitudes instead of all of them. Previous studies have proved that the Chi-Square distance can solve this problem naturally \cite{7,34,55}. 

Besides $MGHMIs$, two moment invariants are chosen for comparison. (1) $GPDs$: $21$ geometric moment invariants of color images proposed in \cite{39}. They have invariance to both the 2D affine transform of spatial coordinates and the 3D diagonal transform of color space. As mentioned previously, this transform model is a kind of restricted $TA$. (2) $SCAMIs$: $24$ geometric moment invariants of color images to $TA$, which were proposed in \cite{34}. From what we know, $GPDs$ and $SCAMIs$ are the only two kinds of existing moment invariants, which are invariant simultaneously to viewpoint change and illumination change. 

The classification accuracy rates from these invariants are listed in the first column of Table \ref{table:4}. For any $\sigma\in\{30,40,...,80\}$, twelve normalized $MGHMIs$ achieve $100\%$ classification accuracy. Using $SCAMIs$, we derive the same result because they are also invariant to $RA$ in theory. The rate $95.83\%$ from $21$ $GPDs$ is slightly lower than $MGHMIs$ and $SCAMIs$, because they are just invariant to the 3D diagonal transform of color space.

\begin{table*}
	\caption{\label{table:4} The classification accuracy from various moment invariants on RGB images disturbed by Gaussian noise.}
	\centering
	\begin{tabular}{p{2.8cm}ccccccc}
		\toprule[1.3pt]
		 \makecell[l]{$\sigma_{G}$\\SNR}&\makecell[c]{$0.000$\\$/$}&\makecell[c]{$0.005$\\22.24 dB}&
		 \makecell[c]{$0.010$\\12.74 dB}&\makecell[c]{$0.015$\\11.09 dB}&
		 \makecell[c]{$0.020$\\9.93 dB}&\makecell[c]{$0.025$\\9.05 dB}&
		 \makecell[c]{$0.030$\\8.34 dB}\\
		\toprule[1.1pt]
		$GPDs$ \cite{39} & 95.83 & 94.17 & 51.67 & 37.00 & 28.17 & 19.33 & 15.00\\
		$SCAMIs$ \cite{34} & \textbf{100.00} & 64.83 & 42.50 & 34.17 & 29.50 & 26.67 & 19.33\\
		\midrule
		$MGHMIs (\sigma=30)$ & \textbf{100.00} & 92.83 & 87.33 & 84.83 & 82.83 & 82.33 & 76.83\\
	    $MGHMIs (\sigma=40)$ & \textbf{100.00} & 94.67 & 91.67 & 89.00 & 89.33 & 85.00 & 83.83\\
	    $MGHMIs (\sigma=50)$ & \textbf{100.00} & 98.67 & 98.00 & 97.00 & \textbf{97.50} & 95.33 & \textbf{95.83}\\
		$MGHMIs (\sigma=60)$ & \textbf{100.00} & \textbf{98.83} & \textbf{98.17} & \textbf{97.17} & 96.17 & 94.00 & 94.50\\
		$MGHMIs (\sigma=70)$ & \textbf{100.00} & 98.67 & 97.67 & 96.50 & 95.00 & \textbf{95.50} & 93.50\\
		$MGHMIs (\sigma=80)$ & \textbf{100.00} & 97.67 & 96.33 & 96.00 & 94.83 & 94.00 & 93.17\\
		\midrule
	\end{tabular}
\end{table*}

\begin{table*}
	\caption{\label{table:5} The classification accuracy from various moment invariants on RGB images disturbed by Salt \& Pepper noise.}
	\centering
	\begin{tabular}{p{2.8cm}ccccccc}
		\toprule[1.3pt]
		\makecell[l]{$r$\\SNR}&\makecell[c]{$0.00$\\$/$}&\makecell[c]{$0.02$\\26.99 dB}&
		\makecell[c]{$0.04$\\17.50 dB}&\makecell[c]{$0.06$\\15.79 dB}&
		\makecell[c]{$0.08$\\14.55 dB}&\makecell[c]{$0.10$\\13.60 dB}&
		\makecell[c]{$0.12$\\12.80 dB}\\
		\toprule[1.3pt]
		$GPDs$ \cite{39} & 95.83 & 94.67 & 90.33 & 80.17 & 72.33 & 61.83 & 54.67\\
		$SCAMIs$ \cite{34} & \textbf{100.00} & 82.17 & 49.00 & 45.50 & 44.33 & 44.00 & 40.33\\
		\midrule
		$MGHMIs (\sigma=30)$ & \textbf{100.00} & 94.33 & 94.33 & 92.67 & 90.67 & 90.67 & 88.00\\
		$MGHMIs (\sigma=40)$ & \textbf{100.00} & 95.50 & 94.17 & 93.17 & 93.17 & 92.83 & 92.33\\
		$MGHMIs (\sigma=50)$ & \textbf{100.00} & 99.00 & \textbf{99.00} & \textbf{99.00} & 98.00 & \textbf{99.00} & \textbf{98.67}\\
		$MGHMIs (\sigma=60)$ & \textbf{100.00} & \textbf{99.17} & \textbf{99.00} & \textbf{99.00} & 97.83 & 98.50 & 98.50\\
		$MGHMIs (\sigma=70)$ & \textbf{100.00} & 98.67 & 98.50 & 98.83 & \textbf{98.17} & 98.50 & 98.50\\
		$MGHMIs (\sigma=80)$ & \textbf{100.00} & 97.83 & 98.33 & 98.50 & 97.33 & 98.50 & 98.50\\
		\midrule
	\end{tabular}
\end{table*}

Further, we test the robustness of $MGHMIs$ of RGB images to additive noises. Specifically, we add a zero-mean Gaussian noise to test images and then similarly repeat the classification experiment. For the Gaussian noise, we set the standard deviation $\sigma_{G}=0.005\times K$, where $K=1,2,...,6$. With the increase of $\sigma_{G}$, the signal-to-noise ratio (SNR) gradually decreases from 22.24 dB to 8.34 dB. Fig.\ref{figure:4(b)} (top row) shows different degraded versions of a test image. 

The classification results are listed in Table \ref{table:4}. It is obvious that $MGHMIs$ significantly outperform $SCAMIs$ and $GPDs$. As stated earlier, $SCAMIs$ and $GPDs$ are extremely sensitive with respect to noise because they are constructed using geometric moments. Even if test images are disturbed by low Gaussian noise ($\sigma_{G}=0.005$, SNR=22.24 dB), the success rate from $SCAMIs$ drops from $100\%$ to $64.83\%$. In contrast, $MGHMIs$ are much more robust to Gaussian noise. Even for heavy noise ($\sigma_{G}=0.030$, SNR=8.34 dB), they still achieve $95.83\%$ accuracy rate ($GPDs$: 15.00$\%$, $SCAMIs$: $19.33\%$). Note that when calculating $MGHMIs$ using larger $\sigma$, such as 60 and 70, we can obtain better results. In fact, the Gaussian function involved in $\hat{H}_{p}(x;\sigma)$ naturally smoothes images and reduces the level of noise. As the scale $\sigma$ increases, the degree of smoothing increases. 

As shown in Table \ref{table:5}, we also conducted the classification on test images disturbed by Salt \& Pepper noise (see the bottom row of Fig.\ref{figure:4(b)}), where the noise density $r=0.02\times K$ and $K=1,2,...,6$. The performance of the three features slightly rises, and $MGHMIs$ is also the best. For example, when setting $r=0.12$ (SNR=12.80), the accuracy rates from $SCAMIs$, $GPDs$ and $MGHMIs(\sigma=50)$ are $40.33\%$, $54.67\%$ and $98.67\%$, respectively.       

\begin{figure}
	\centering
	\subfloat[Ten texture classes are selected from Outex-TC-00014 dataset.]
	{\includegraphics[height=31mm,width=82mm]{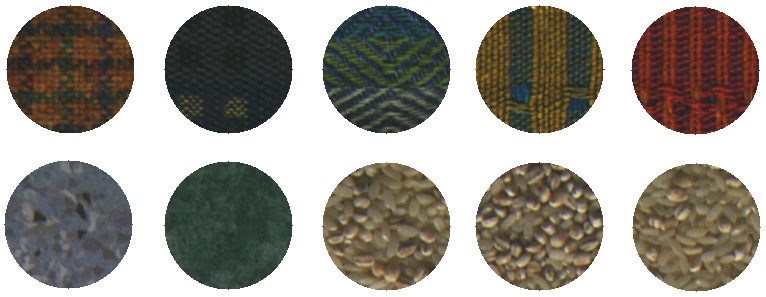}\label{figure:6(a)}\hfill}\\
	\subfloat[Some training images and the corresponding test images. They are acquired under different illumination conditions. ]
	{\includegraphics[height=35mm,width=82mm]{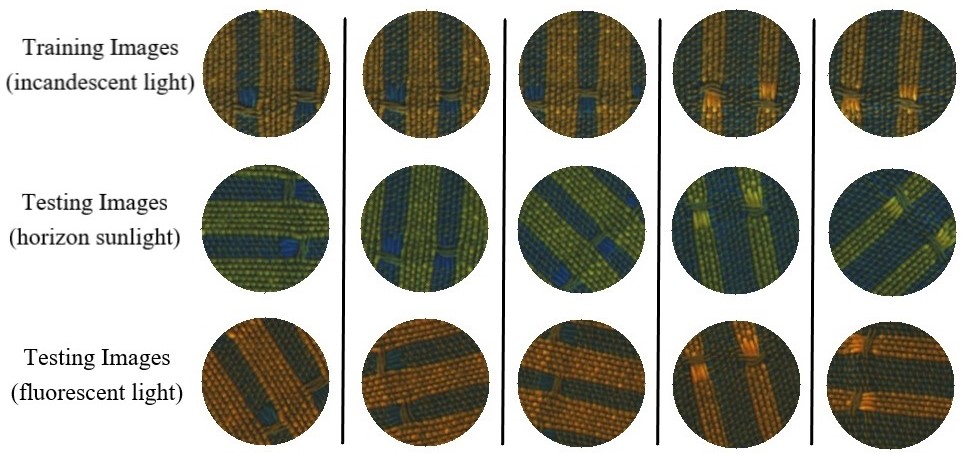}\label{figure:6(b)}\hfill}
	\caption{Real RGB images used to test the performance of $MGHMIs$.}\label{figure:6}
\end{figure}
  
\begin{figure}
 	\centering
 	\includegraphics[height=47mm,width=85mm]{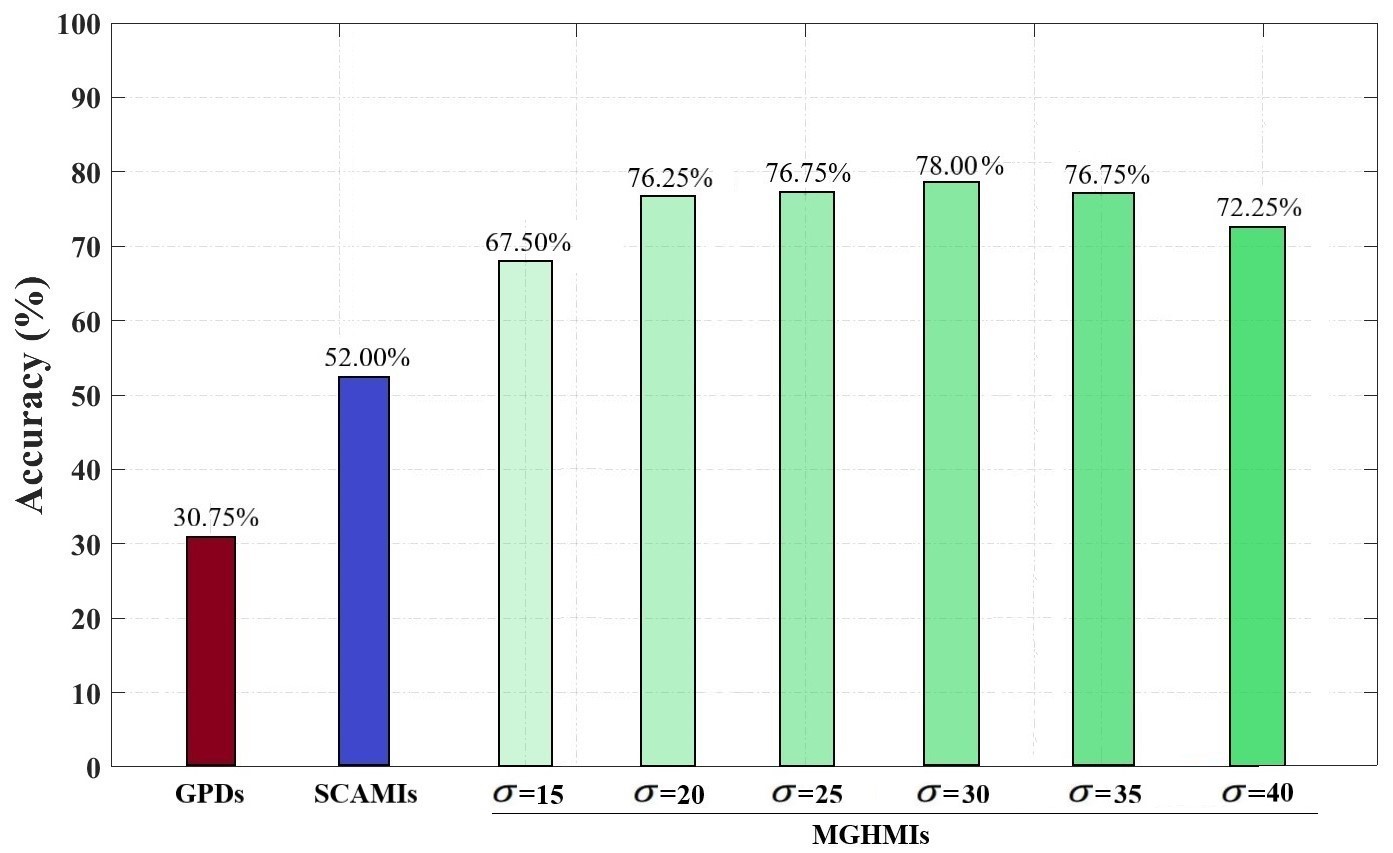}\\
 	\caption{The accuracy rates from various image features on real RGB images.}\label{figure:7}
\end{figure}

Finally, real RGB images are used to evaluate the performance of these normalized $MGHMIs$. Outex-TC-00014 dataset contains $68$ texture classes and provides $20$ texture patches per class \cite{56}. Three $128\times 128$ images are acquired for each texture patch using three different illumination sources, including incandescent light, horizon sunlight and fluorescent light. As shown in Fig.\ref{figure:6(a)}, we select ten texture classes from this dataset. These textures contain different color regions. Then, $10\times 20=200$ images captured using incandescent light are used as training images, while $10\times20\times2=400$ images under the other two light conditions are used as test images. 

Our task is to establish correct matches between $200$ training images and $400$ test images based on feature vectors. In order to increase the difficulty of this task, each test image is further rotated by a random angle. As shown in the first row of Fig.\ref{figure:6(b)}, the training set contains similar texture images from the same class. Thus, finding the correct image from the training set for a given test image is challenging. 

The matching accuracies from $GPDs$, $SCAMIs$ and $MGHMIs$ on $400$ test images are shown in Fig.\ref{figure:7}. We can find that $MGHMIs$ are still the best. For example, when setting $\sigma=30$, the accuracy rate of $78.00\%$ from $MGHMIs$ is significantly higher than the other features $(GDPs:30.75\%,~SCAMIs:52.00\%)$. This result is consistent with the classification results on synthetic RGB images. Also, it proves that the 3D affine transform of RGB space can model realistic color variation caused by illumination change.  

\subsection{MGHMIs of 2D vector fields to TR and RA}
\label{section:5.2}
\begin{figure}
	\centering
	\includegraphics[height=8mm,width=86mm]{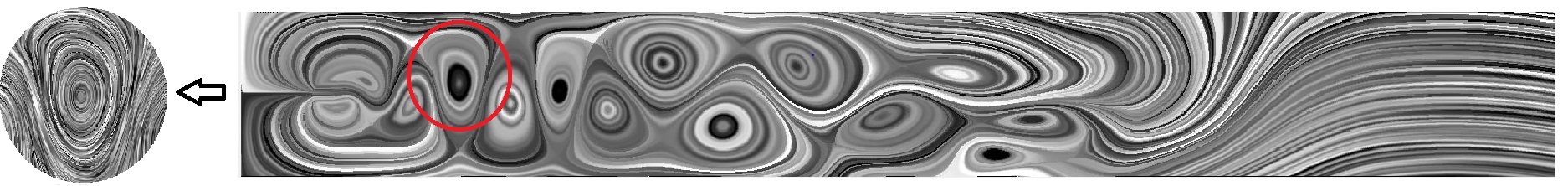}\\
	\caption{Frame 500 with a selected $65 \times 65$ vortex template (red circle).}\label{figure:8}
\end{figure}

As shown in Table \ref{table:2}, for 2D vector fields, we generate seven $MGHMIs$ to $TR$ and seven $MGHMIs$ to $RA$, respectively. In this subsection, they are denoted as $TR\_MGHMIs$ and $RA\_MGHMIs$. We mainly test their applicability in vortex detection, which is an essential task in the field of fluid dynamics engineering. For this purpose, we first download a 1501-frame video of 2D cylinder flow from \url{https://cgl.ethz.ch/Downloads/Data/ScientificData/cylinder2d_vti.zip}, which shows the time-development of a von K$\acute{a}$rm$\acute{a}$n vortex street. In each video frame, there are many similar patterns of swirling vortices with different orientations. Fig.\ref{figure:8} shows the line integral convolution (LIC) \cite{57} of frame 500. The spatial resolution of all frames is $80\times640$. 

\begin{figure}
	\centering
	\subfloat[Some templates randomly selected from frame 500.]
	{\includegraphics[height=13mm,width=75mm]{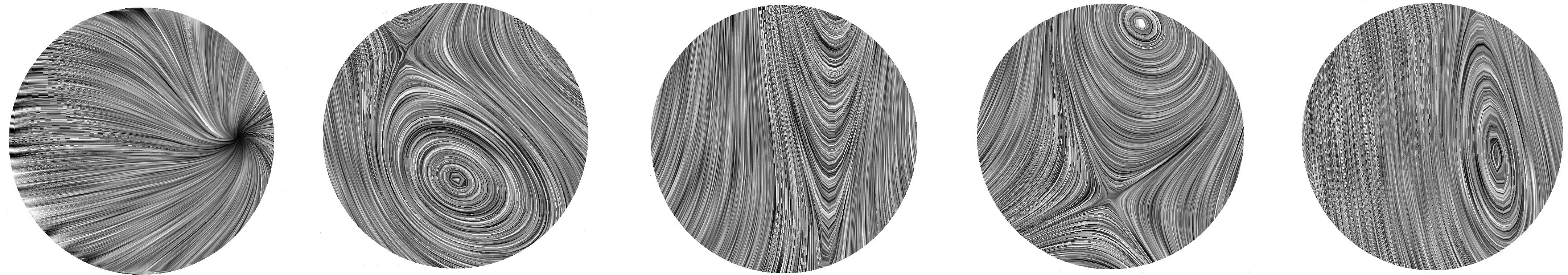}\label{figure:9(a)}\hfill}\\
	\subfloat[Four special $TR$ versions of a template.]
	{\includegraphics[height=13mm,width=75mm]{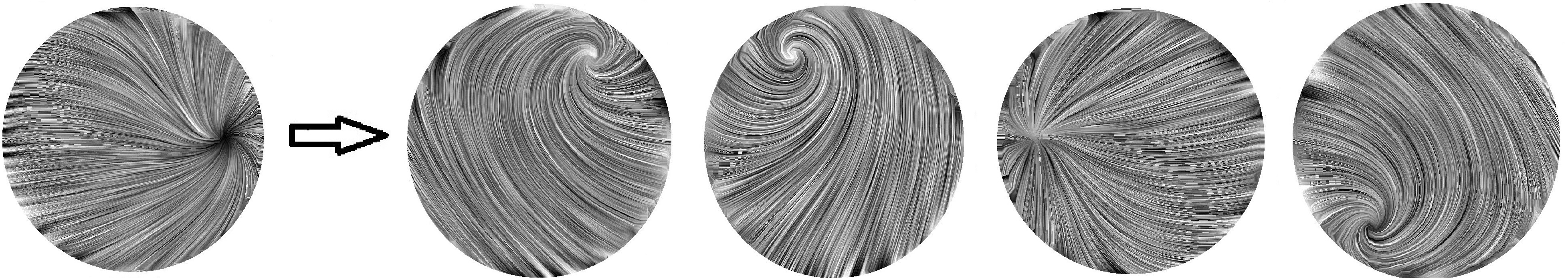}\label{figure:9(b)}\hfill}
	\caption{Some examples of 2D vector fields used for testing the stability of moment invariants.}\label{figure:9}
\end{figure}

\begin{figure*}
	\centering
	\subfloat[$sTR\_GHMIs$.]
	{\includegraphics[height=40mm,width=54mm]{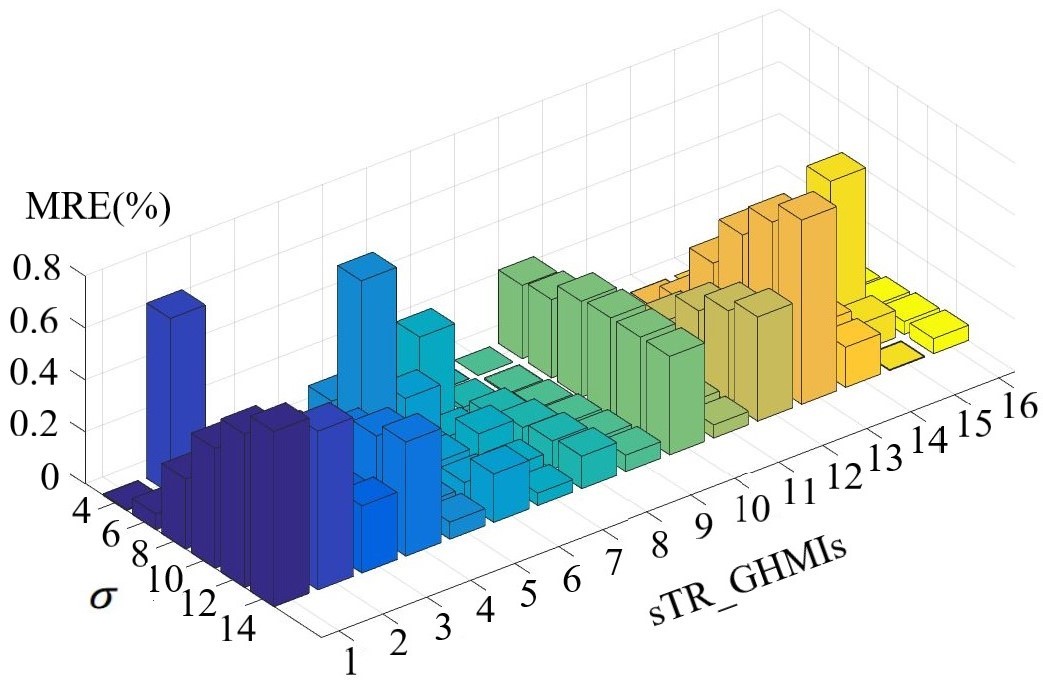}\label{figure:10(a)}\hfill}~~~~
	\subfloat[$TR\_MGHMIs$.]
	{\includegraphics[height=40mm,width=52mm]{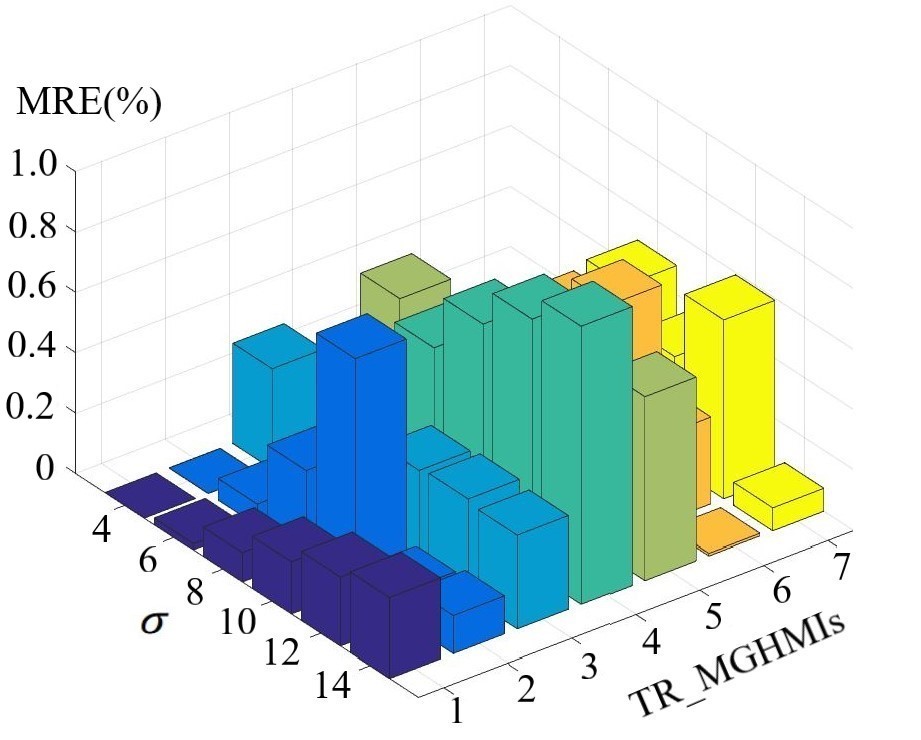}\label{figure:10(b)}\hfill}~~~~
	\subfloat[$RA\_MGHMIs$.]
	{\includegraphics[height=40mm,width=54mm]{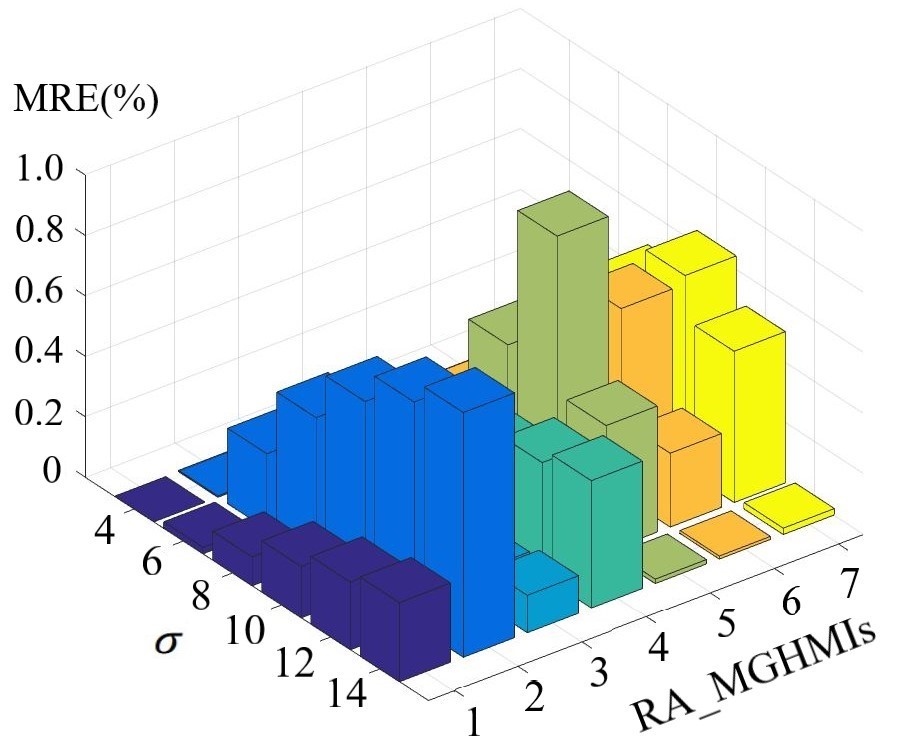}\label{figure:10(c)}\hfill}
	\caption{The $MRE$ of Gaussian-Hermite moment invariants to special $TR$, $TR$ and $RA$.}\label{figure:10}
\end{figure*}

Using the procedure stated in Section \ref{section:5.1}, we first evaluate the stability of $TR\_MGHMIs$ and $RA\_MGHMIs$. First, ten $41\times 41$ templates are randomly selected from frame $500$ (see Fig.\ref{figure:9(a)}). As mentioned in Section \ref{section:2.2}, previous researchers usually model realistic local deformations of vector fields as special $TR$. In special $TR$, $OT=R_{out}$ and $IT=R_{in}$ are identical rotation matrices. Hence, for each template, sixty special $TR$ versions are generated. We set $\theta_{out}=\theta_{in}=1\cdot2\pi/60,2\cdot2\pi/60,\cdots, 60\cdot2\pi/60$, where $\theta_{out}$ and $\theta_{in}$ represent rotation angles of $R_{out}$ and $R_{in}$, respectively. Several transformed examples of a template are shown in Fig.\ref{figure:9(b)}. Since special $TR\subset TR \subset RA$, $TR\_MGHMIs$ and $RA\_MGHMIs$ are also invariant to special $TR$. 

In \cite{48}, Yang et al. constructed Gaussian-Hermite moment invariants of 2D vector fields, which are only invariant to special $TR$. In our paper, they are denoted as $sTR\_GHMIs$. For comparison, we also generate eight independent $sTR\_GHMIs$ up to the third order. The value of each $sTR\_GHMIs$ is a complex number containing a real and imaginary parts. This means that we actually get $16$ invariant values from eight $sTR\_GHMIs$. 

When setting the scale parameter $\sigma=4,6,...,14$ ($4\approx 41/9$ and $14\approx 41/3$), the $MRE$ of $sTR\_GHMIs$, $TR\_MGHMIs$ and $RA\_MGHMIs$ can be seen in Fig.\ref{figure:10}. We can observe that all of them are less than $1.0\%$, which confirms the invariance of these moment invariants to special $TR$. 

As stated in Section \ref{section:2.2}, in practical cases, vector values will be disturbed by different factors, such as random noise, which results in that $R_{out}$ and $R_{in}$ are not precisely equal. This will reduce the stability and discriminability of $sTR\_GHMIs$. To verify this, we test the performance of $sTR\_GHMIs$ and $TR\_MGHMIs$ on the classification task. Ten templates are used as training data, while $600$ special $TR$ versions of them are used as test data. Similar to Section \ref{section:5.1}, we still utilize the Nearest Neighbor classifier based on the Chi-Square distance. 

\begin{figure}
	\centering
	\subfloat[Zero-mean Gaussian noise, where $\sigma_{G}$ represents its standard deviation.]
	{\includegraphics[height=30mm,width=40mm]{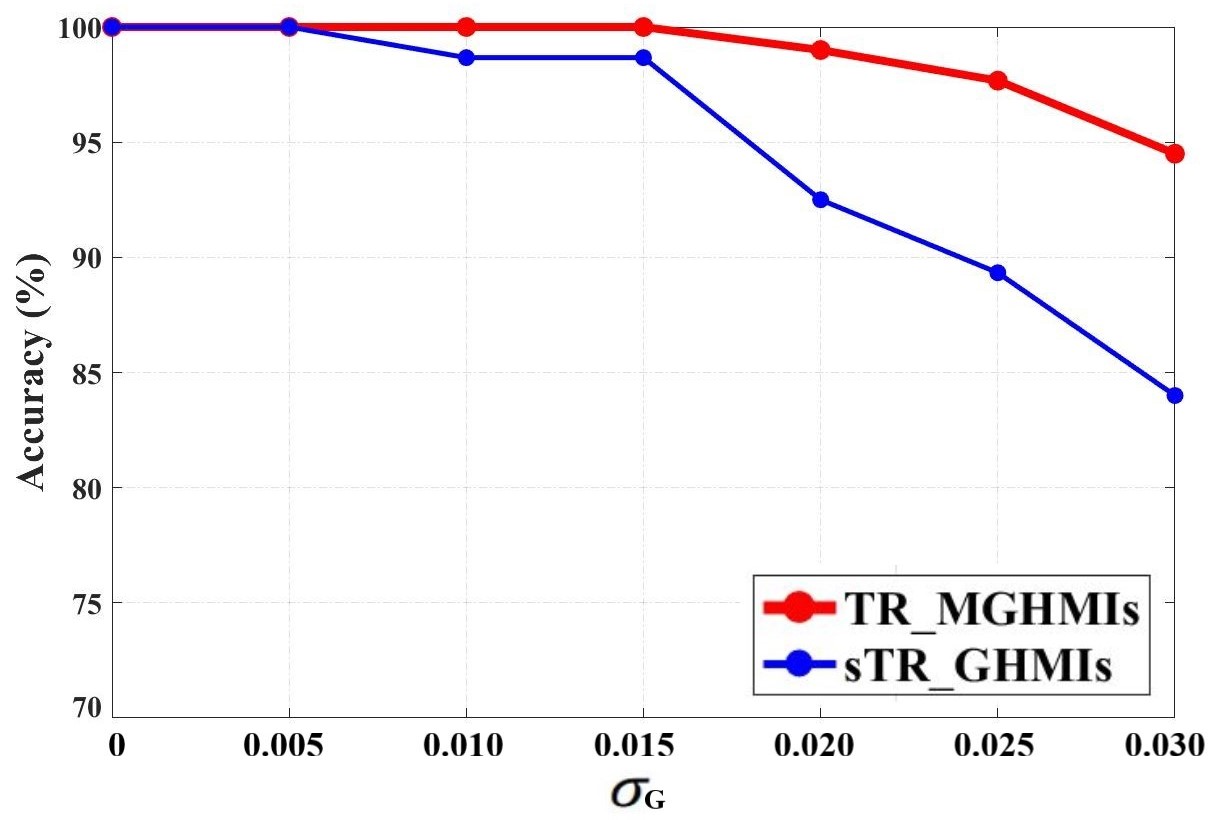}\label{figure:11(a)}\hfill}~~
	\subfloat[Angle error between $R_{out}$ and $R_{in}$.]
	{\includegraphics[height=30mm,width=40mm]{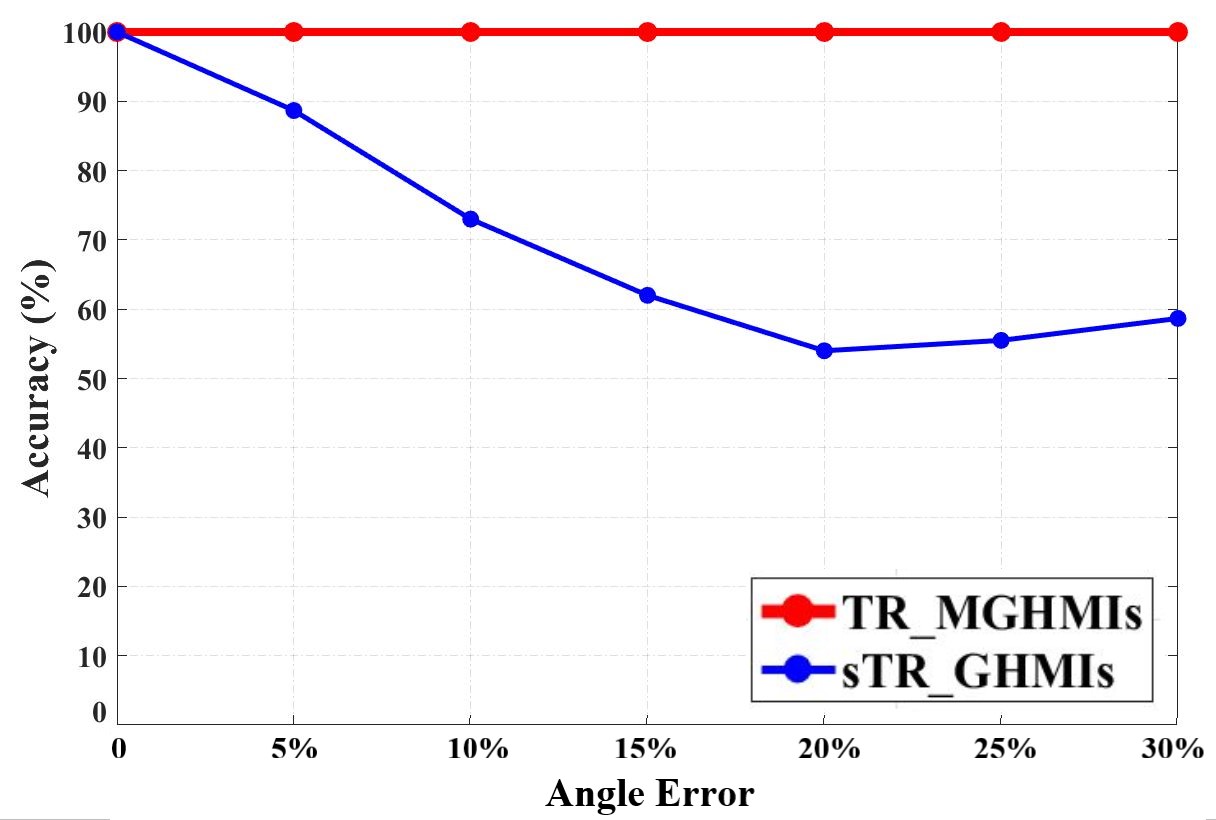}\label{figure:11(b)}\hfill}
	\caption{The classification accuracies from $sTR\_GHMIs$ and $TR\_MGHMIs$ on 2D vector fields disturbed by different factors.}\label{figure:11}
\end{figure}

Fig.\ref{figure:11(a)} shows the accuracy rates from $sTR\_GHMIs$ and $TR\_MGHMIs$ on test data disturbed by different Gaussian noise levels. It can be seen that, although both $sTR\_GHMIs$ and $TR\_MGHMIs$ are constructed using Gaussian-Hermite moments up to the same order, the noise robustness of $TR\_MGHMIs$ is better than $sTR\_GHMIs$. For example, when the standard deviation of Gaussian noise is $0.030$, $TR\_MGHMIs$ correctly classify $94.50\%$ test data while the accuracy rate from $sTR\_GHMIs$ drops to $84.00\%$. This is because Gaussian noise makes $R_{out}$ imprecisely equal to $R_{in}$, resulting in the poor performance of $sTR\_GHMIs$. 

An interesting question is how much angle error between $R_{out}$ and $R_{in}$ can be tolerated by $sTR\_GHMIs$. To this end, we make $\theta_{out}=\theta_{in}+\epsilon\cdot\theta_{in}$ and regenerate test data again, where $\epsilon=0\%,5\%,10\%,\cdots,30\%$. Then, the performance of $sTR\_GHMIs$ and $TR\_MGHMIs$ is evaluated on these test data with angle error. As shown in Fig.\ref{figure:11(b)}, $TR\_MGHMIs$ always maintain $100\%$ accuracy rate as angle error increases because they are invariant to general $TR$. In contrast, even if there is only $10\%$ difference between $\theta_{out}$ and $\theta_{in}$, the performance of $sTR\_GHMIs$ decreases from $100\%$ to $73.00\%$.  

Then, our task is to find all vortices in frames 500 $\sim$ 1500 of the cylinder flow video. We select a $65 \times 65$ template with a typical vortex from frame 500 (see Fig.\ref{figure:8}), and calculate $sTR\_GHMIs$, $TR\_GHMIs$ and $RA\_GHMIs$ on the template by setting the scale parameter $\sigma=17$ ($17\approx 65/4$). In a similar fashion, the values of these moment invariants are also calculated on a $65 \times 65$ neighborhood of each point in 1000 frames. Again, we use the Chi-Square distance to measure the similarity between a given point and the template in the space of features. 

In previous papers \cite{47,48,49,50}, all local minima of feature distance below a user-defined threshold were regarded as possible locations of vortices in a frame. However, this criterion discards locations with smaller distances (more similar to the template) but do not meet local minimum conditions. This makes the detection results unable to fully reflect the stability and discriminability of moment invariants. To resolve this issue, in this paper, we sort all points of a frame $(80\times640=51200)$ in ascending order by the Chi-Square distance and then mark the first 2000 points on the frame. 

Recently, the paper \cite{50} proposed moment invariants of 2D vector fields to $TA$, denoted as $TA\_GMIs$, and derived eight $TA\_GMIs$ up to the third order, which form an independent set. We also calculate these eight $TA\_GMI$ for comparison. Recall again that the relationship between four transform models is special $TR\subset TR \subset RA \subset TA$.        

\begin{figure*}
	\centering
	\subfloat[$sTR\_GHMIs$ \cite{48}]
	{\includegraphics[height=8mm,width=80mm]{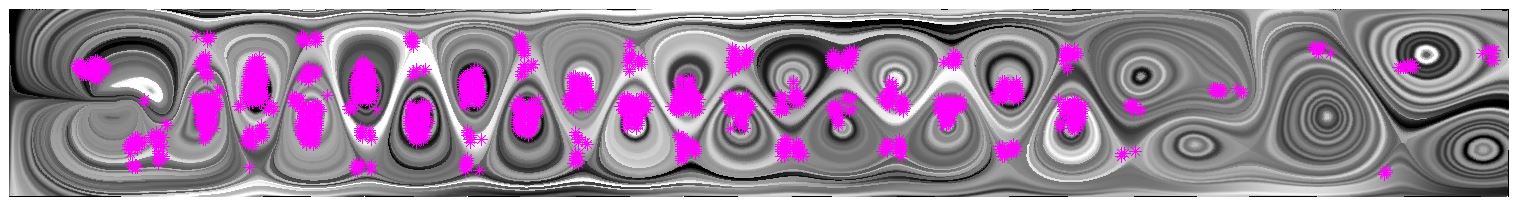}\label{figure:12(a)}\hfill}~~~~
	\subfloat[$TR\_MGMIs$]
	{\includegraphics[height=8mm,width=80mm]{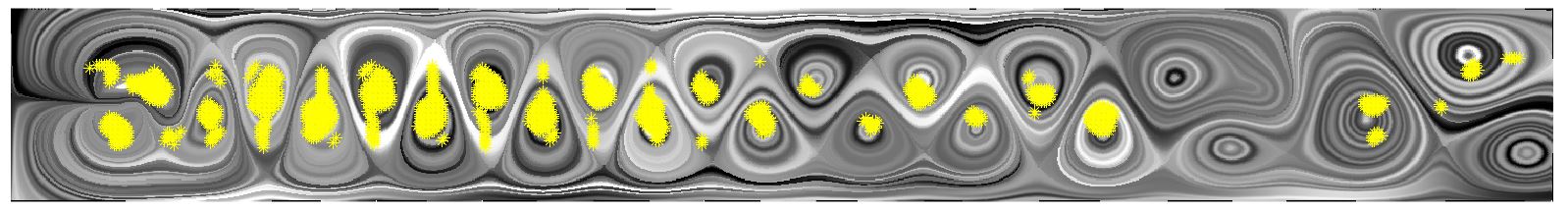}\label{figure:12(b)}\hfill}\\
	\subfloat[$RA\_MGHMIs$]
	{\includegraphics[height=8mm,width=80mm]{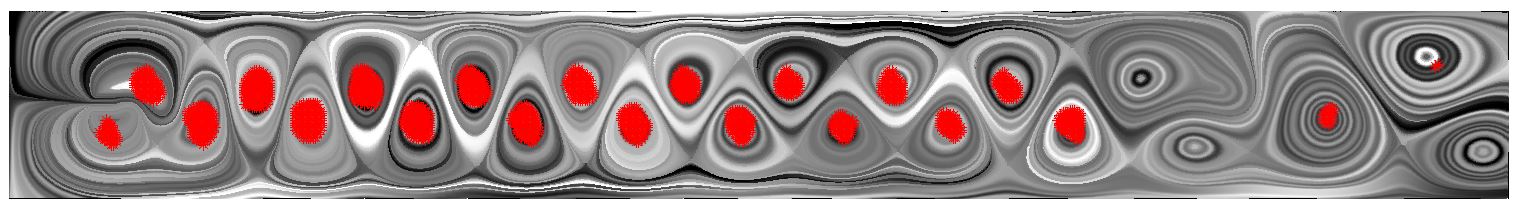}\label{figure:12(c)}\hfill}~~~~
	\subfloat[$TA\_GMIs$ \cite{50}]
	{\includegraphics[height=8mm,width=80mm]{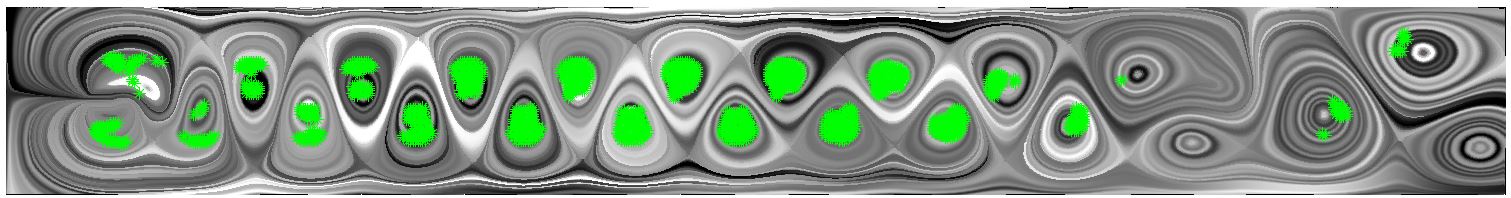}\label{figure:12(d)}\hfill}
	\caption{Detected locations from various moment invariants on frame 1000.}\label{figure:12}
\end{figure*}

\begin{figure*}
	\centering
	\subfloat[$sTR\_GHMIs$ \cite{48}]
	{\includegraphics[height=8mm,width=80mm]{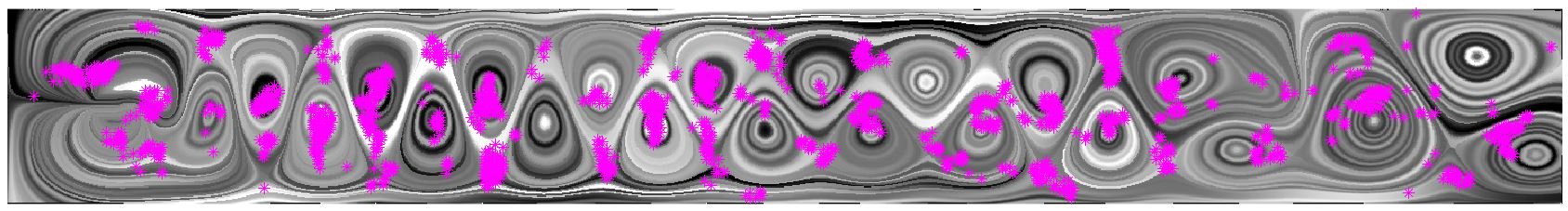}\label{figure:13(a)}\hfill}~~~~
	\subfloat[$TR\_MGMIs$]
	{\includegraphics[height=8mm,width=80mm]{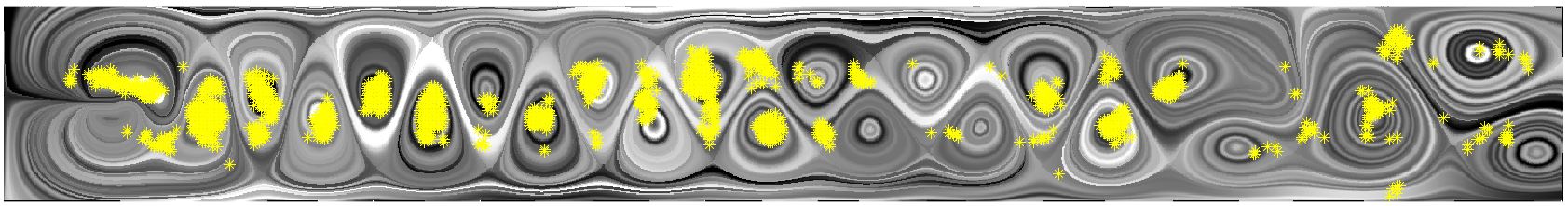}\label{figure:13(b)}\hfill}\\
	\subfloat[$RA\_MGHMIs$]
	{\includegraphics[height=8mm,width=80mm]{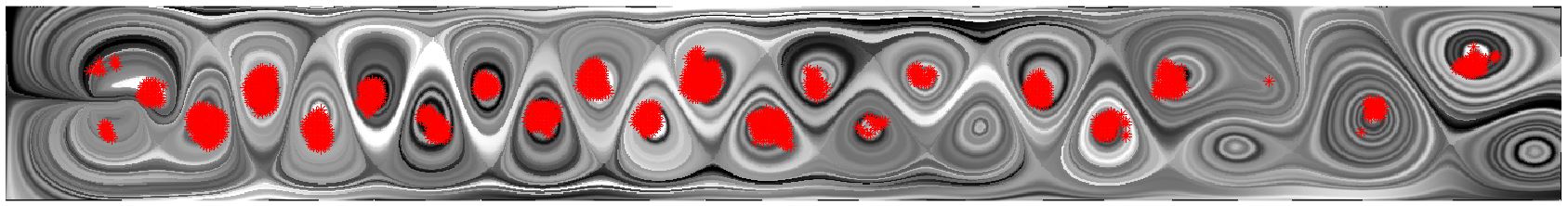}\label{figure:13(c)}\hfill}~~~~
	\subfloat[$TA\_GMIs$ \cite{50}]
	{\includegraphics[height=8mm,width=80mm]{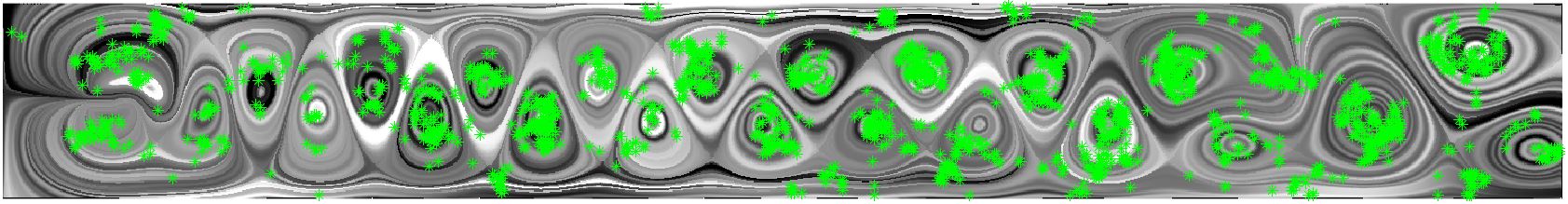}\label{figure:13(d)}\hfill}
	\caption{Detected locations from various moment invariants on a noisy version of frame 1000.}\label{figure:13}
\end{figure*}

It isn't easy to evaluate detection results quantitatively because we do not know the ground-truth locations of vortices on each frame. Fortunately, previous researchers have found that visual inspection of resulting videos provides a good insight into the performance of features \cite{47,48,49,50}. The videos showing detected locations from various moment invariants can be downloaded from the link \url{https://drive.google.com/drive/folders/1mJjv15kMwuyTKGmCGN9nlySyOVz14q9m?usp=share_link}. 

As an instance, detected results on frame 1000 are shown in Fig.\ref{figure:12}. First, it can be seen that $TR\_MGHMIs$, $RA\_MGHMIs$ and $TA\_GMIs$ correctly locate most of the vortices and outperform $sTR\_GHMIs$. In fact, similar vortices in a frame are not special $TR$ versions of the same template. They are related to each other by a more complex transform model, which means that $OT$ does not need to be exactly equal to $IT$. Secondly, $RA\_MGHMIs$ and $TA\_GMIs$ perform better than $TR\_MGHMIs$. This indicates that affine transform can better describe realistic deformations acting on vector values than simple rotation. Thirdly, we do not find that $TA\_GMIs$ achieve better detected results than $RA\_MGHMIs$. Compared with $RA$, $TA$ generalizes $IT$ from spatial rotation $R_{in}$ to affine transform $A_{in}$. Thus, in theory, $TA\_GMIs$ can detect those vortices with different shapes. For example, in frame 1000, several rightmost vortices can be regarded as $TA$ versions of the template because they have different scales. $RA$ cannot describe spatial scale change. However, like the other invariants, $TA\_GMIs$ also fail to detect these vortices (Fig.\ref{figure:12(d)}). The main reason is that all moment invariants are calculated on a circular region with a fixed radius $(65)$ around each location. As stated in Section \ref{section:2.2}, to comply with the definition of $TA$, we should have first transformed a circular region into an elliptical region using $IT=A_{in}$ and then calculated $TA\_GMIs$ on this affine-equivariant region. Unfortunately, this operation is almost impossible in practice because we don't know the parameters of $A_{in}$. That's why our paper demands $IT$ must be spatial rotation $R_{in}$ instead of affine transform $A_{in}$.

To test the robustness to noise, we add a zero-mean Gaussian noise to each frame, where the average SNR on all frames is -8.780 dB, and then repeat the detection experiment. The resulting videos can also be found at the above link. The visual results on noisy frame 1000 are shown in Fig.\ref{figure:13}. By comparing Fig.\ref{figure:12} and \ref{figure:13}, we can see that the performance of $TA\_GMIs$ decreases clearly. This is because they are constructed using geometric moments sensitive to random noise. In addition, although $sTR\_GHMIs$, $TR\_MGHMIs$ and $RA\_MGHMIs$ are all constructed using Gaussian-Hermite moments up to the third order, the robustness of $RA\_MGHMIs$ significantly outperforms the other two and achieve an excellent detection result on noisy data. The invariance of $RA\_MGHMIs$ to more complex $OT$ gives them better numerical stability when vector values are corrupted and distorted by deformations and noise. It can be understood that in some cases, random noise will change $OT$ from a rotation matrix to a more complex transformation, and at this time, the affine transform model is more suitable to describe it. 

\begin{figure}
	\centering
	\includegraphics[height=40mm,width=80mm]{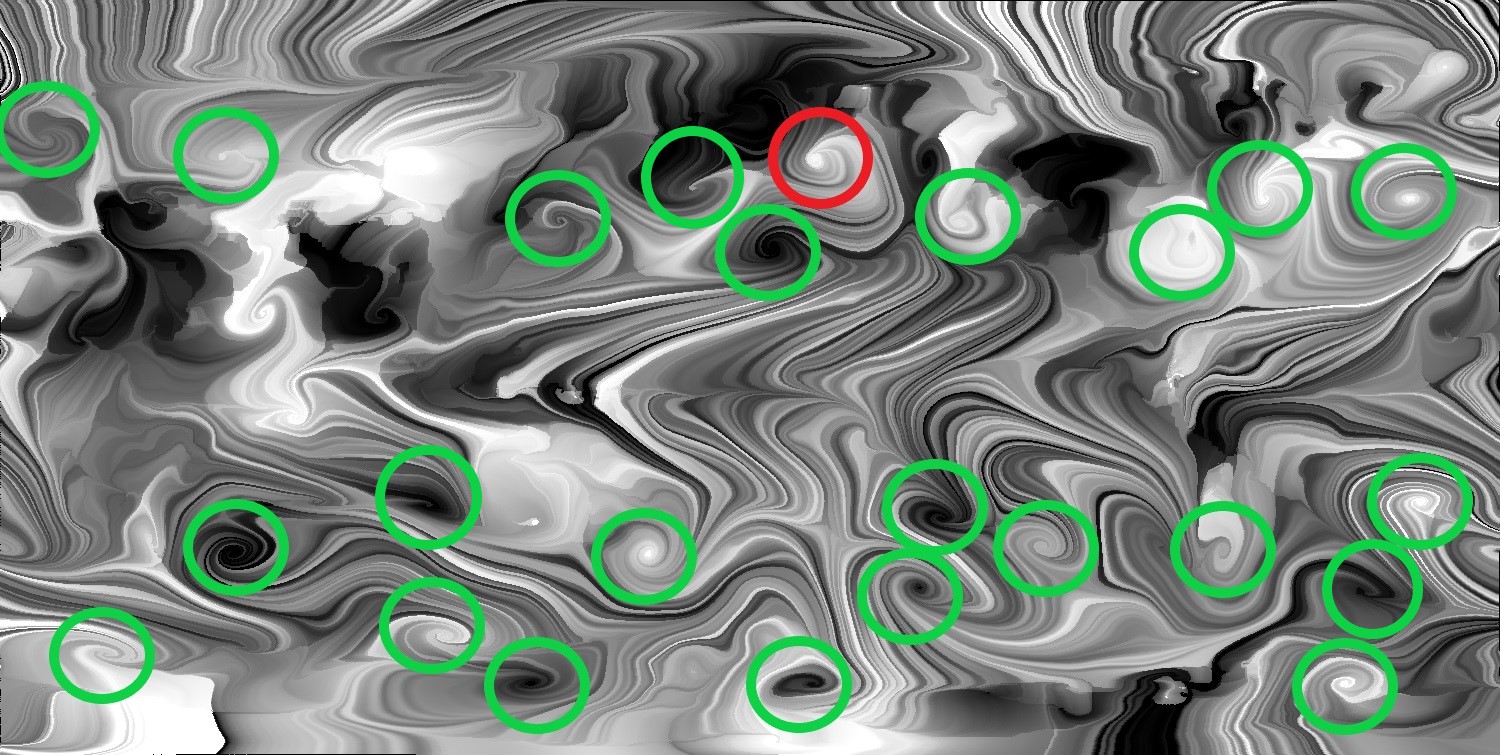}\\
	\caption{A real wind field with one selected $65 \times 65$ vortex template (red circle) and many other vortices with different shapes (green circles).}\label{figure:14}
\end{figure}

\begin{figure*}
	\centering
	\subfloat[$sTR\_GHMIs$ \cite{48}]
	{\includegraphics[height=40mm,width=75mm]{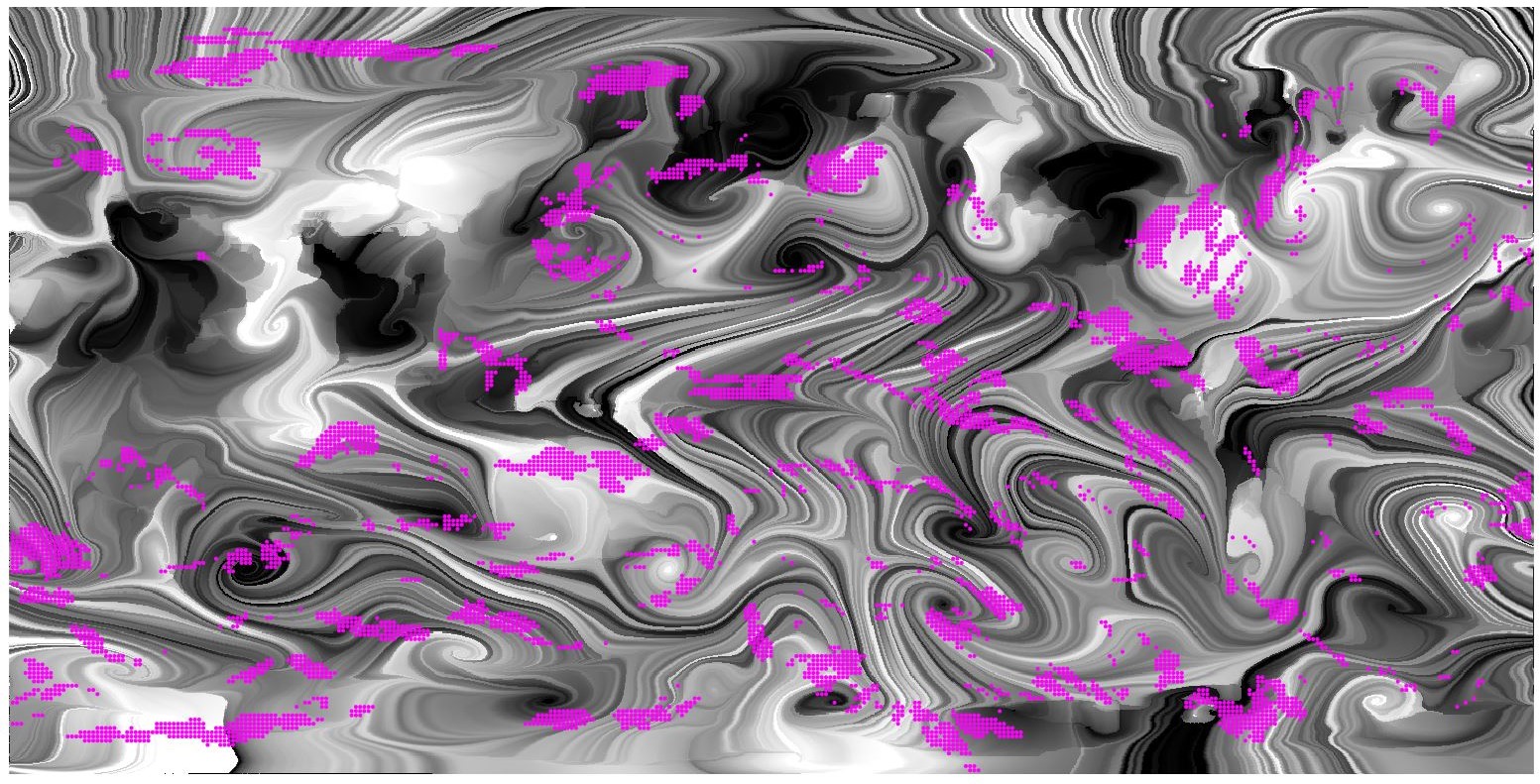}\label{figure:15(a)}\hfill}~~~~~~~~~~~~
	\subfloat[$TR\_MGMIs$]
	{\includegraphics[height=40mm,width=75mm]{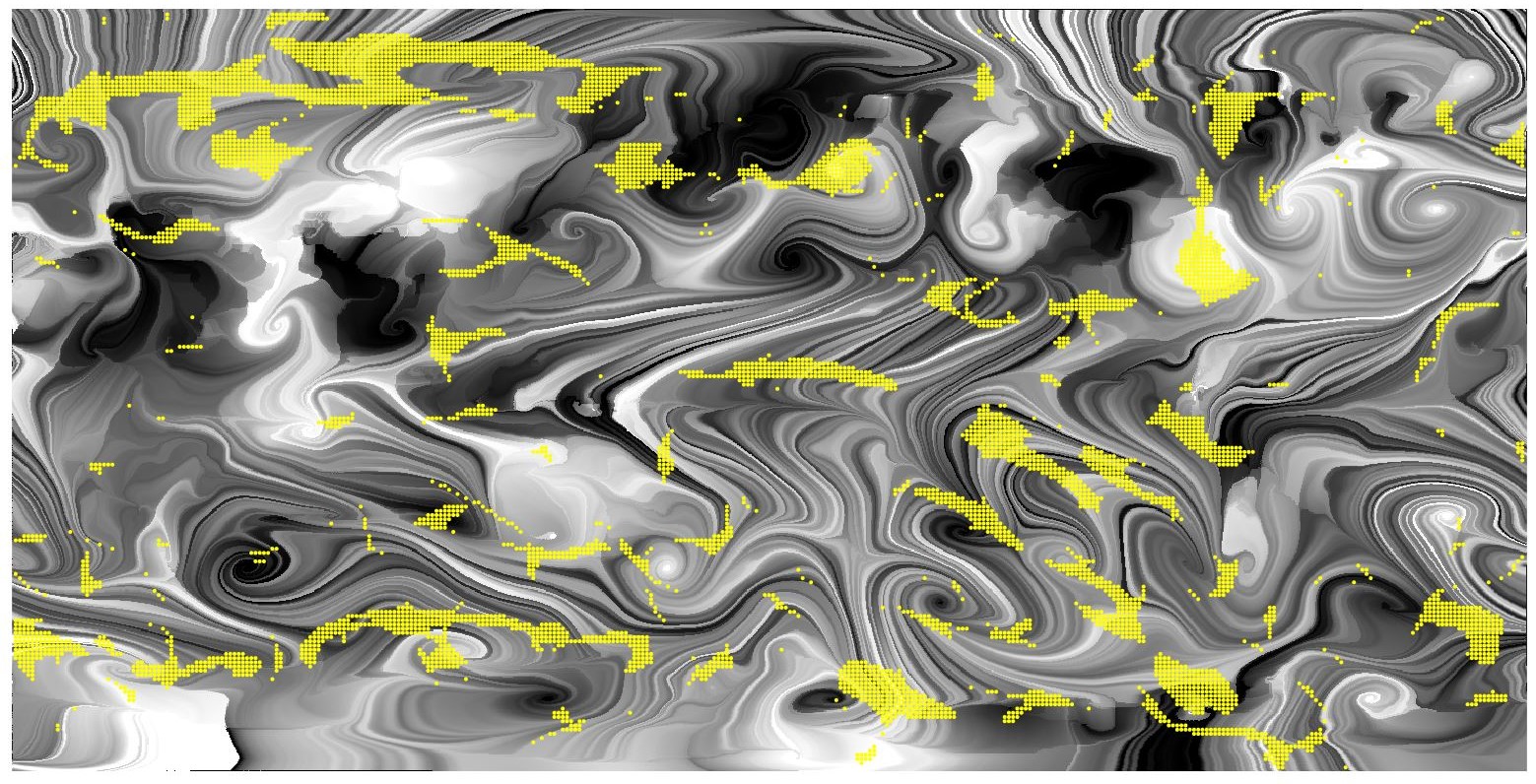}\label{figure:15(b)}\hfill}\\
	\subfloat[$RA\_MGHMIs$]
	{\includegraphics[height=40mm,width=75mm]{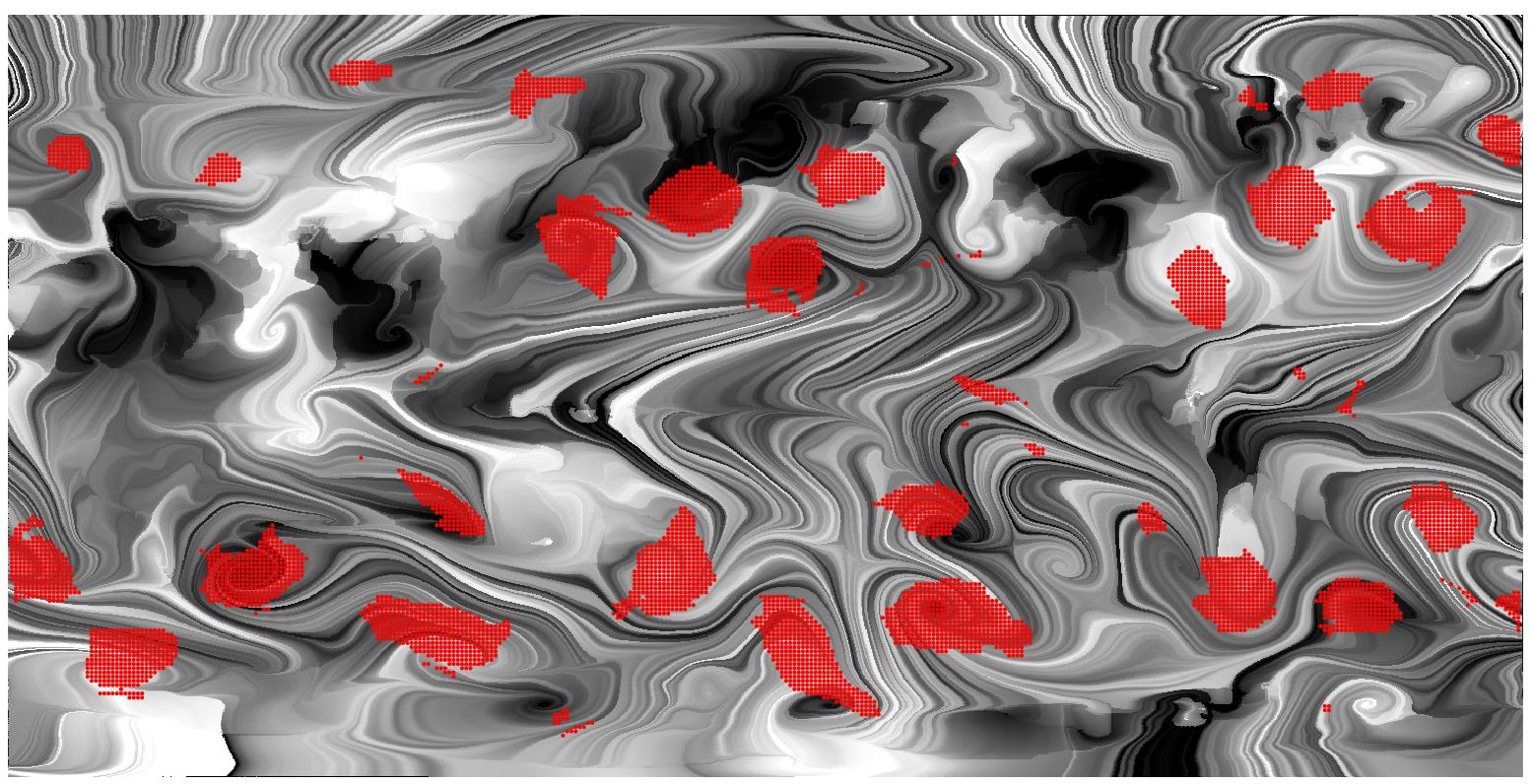}\label{figure:15(c)}\hfill}~~~~~~~~~~~~
	\subfloat[$TA\_GMIs$ \cite{50}]
	{\includegraphics[height=40mm,width=75mm]{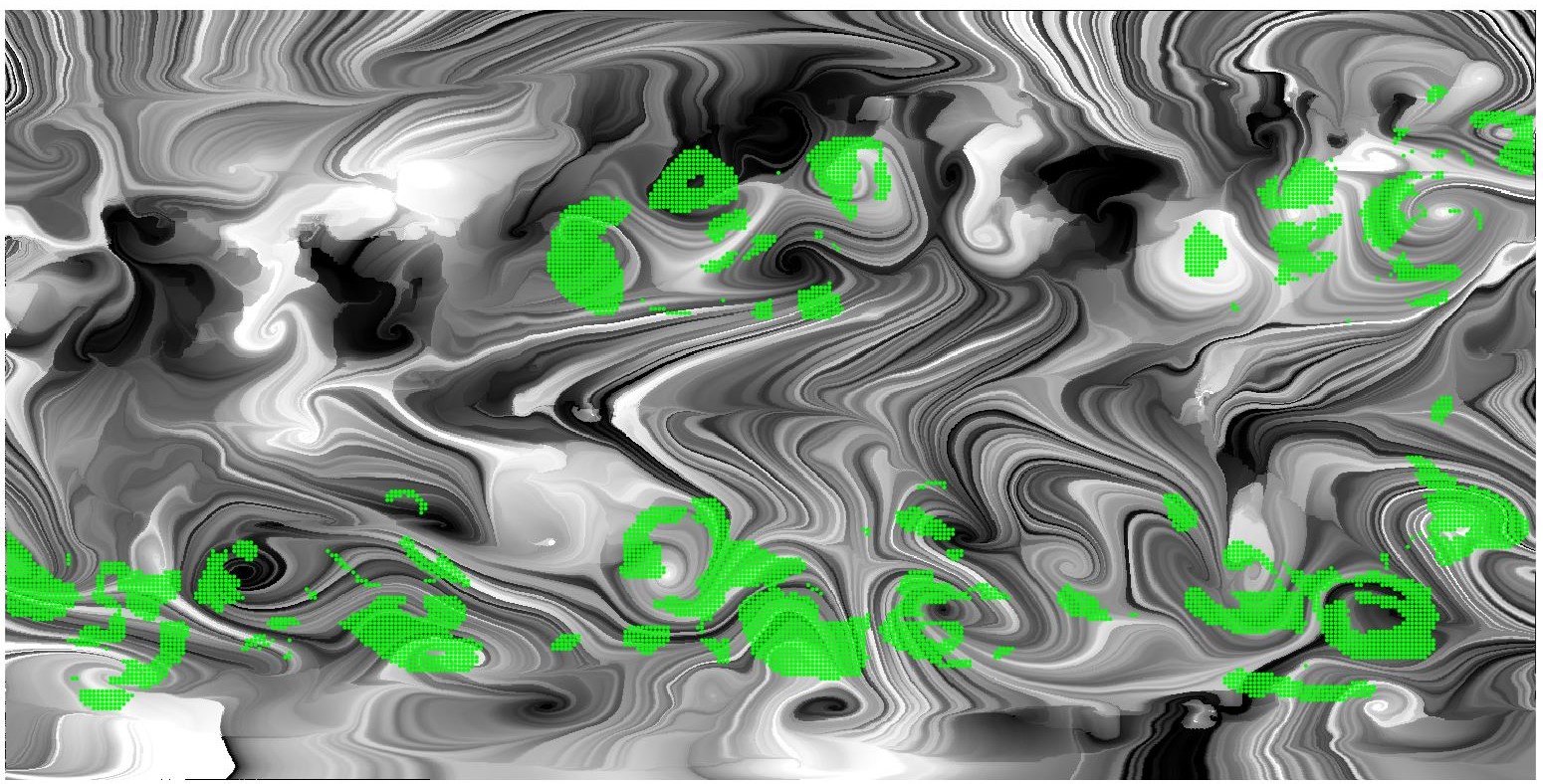}\label{figure:15(d)}\hfill}
	\caption{Detected locations from various moment invariants on a real vector wind field.}\label{figure:15}
\end{figure*}

Similar to \cite{50}, $MGHMIs$ are further used to detect vortices in real 2D wind fields for one month (31 days). These vector wind fields are collected using NOAA satellite and can be downloaded from \url{https://psl.noaa.gov/cgi-bin/data/composites/plot20thc.day.v2.pl}. Their spatial resolution is $181\times 360$. Similar to the above detection experiment, we select a vortex template from a wind field (the 22nd day), calculate various moment invariants, and finally display 6000 locations in each wind field with smaller feature distances from the template. Fig.\ref{figure:14} shows the LIC of the wind field on the 22nd day, the template we selected (red circle), and all vortices we want to detect (green circles). 

The detected results from various moment invariants on this wind field are shown in Fig.\ref{figure:15}. It can be observed that $RA\_MGHMIs$ correctly detect most of the vortices while $sTR\_GHMIs$ and even $TR\_MGHMIs$ perform very poorly. Compared with 2D cylinder flow, vortices in real wind fields are more varied, which means the transformations between them are more complex. In addition, $TA\_GMIs$ perform better than $sTR\_GHMIs$ and $TR\_MGHMIs$ but far worse than $RA\_MGHMIs$. One possible reason is that real data contain random noise. Also, we cannot calculate these $TA\_GMIs$ on affine-equivariant regions. The videos of detected results on all wind fields can also be downloaded from the above link.

\section{Conclusions}
\label{section:6}
In this paper, we propose a unified framework for deriving Gaussian-Hermite moment invariants of general multi-channel functions to rotation-affine transform and total rotation transform and introduce how to utilize the framework to generate all possible $MGHMIs$ for a specific type of multi-channel data. For RGB images, 2D vector fields and color volume data, independent sets of $MGHMIs$ up to low degree and low order are generated. We conduct extensive numerical experiments on synthetic and real multi-channel data to verify the stability and discriminability of $MGHMIs$ and their robustness to additive noise. For comparison, we select nearly all of the moment invariants of multi-channel functions published in previous papers. Our results clearly show that $MGHMIs$ have better performance in RGB image classification and vortex detection. In the future, we plan to combine $MGHMIs$ with deep learning methods and further explore their application in computer vision, pattern recognition and visualization.    

\ifCLASSOPTIONcompsoc
  \section*{Acknowledgments}
\else

  \section*{Acknowledgment}
\fi
This work has partly been funded by the National Key R$\&$D Program of China (No. 2017YFB1002703), the National Natural Science Foundation of China (Grant No. 60873164, 61227802 and 61379082) and the Academy of Finland for Academy Professor project EmotionAI (Grant No. 336116).

\ifCLASSOPTIONcaptionsoff
  \newpage
\fi




\end{document}